\documentclass[10pt,twocolumn,letterpaper]{article}

\usepackage{cvpr}              

\usepackage{graphicx}
\usepackage{amsmath}
\usepackage{amssymb}
\usepackage{booktabs}
\usepackage{multirow}
\usepackage{tikz}
\usepackage{siunitx}
\usepackage{array}
\usepackage{algorithm}
\usepackage{algpseudocode}
\usepackage{xcolor}

\definecolor{cvprblue}{rgb}{0.21,0.49,0.74}
\usepackage[pagebackref,breaklinks,colorlinks,allcolors=cvprblue]{hyperref}

\def\paperID{7618} 
\def\confName{CVPR}
\def\confYear{2025}

\newcommand{\tabincell}[2]{\begin{tabular}{@{}#1@{}}#2\end{tabular}} 

\title{\textit{Early-Bird Diffusion}: Investigating and Leveraging Timestep-Aware Early-Bird Tickets in Diffusion Models for Efficient Training}

\author{Lexington Whalen\thanks{Equal contribution.} , Zhenbang Du$^{*}$, Haoran You$^{*}$, Chaojian Li, Sixu Li,\\
Yingyan (Celine) Lin\\
Georgia Institute of Technology\\
{\tt\small \{lwhalen7,zdu89,hyou37,cli851,sli941,celine.lin\}@gatech.edu}
}

\begin{document}
\maketitle

\begin{abstract}

Training diffusion models (DMs) requires substantial computational resources due to multiple forward and backward passes across numerous timesteps, motivating research into efficient training techniques. In this paper, we propose \textbf{EB-Diff-Train}, a new efficient DM training approach that is orthogonal to other methods of accelerating DM training, by investigating and leveraging Early-Bird (EB) tickets—sparse subnetworks that manifest early in the training process and maintain high generation quality. 
We first investigate the existence of traditional EB tickets in DMs, enabling competitive generation quality without fully training a dense model. 
Then, we delve into the concept of diffusion-dedicated EB tickets, drawing on insights from varying importance of different timestep regions. These tickets adapt their sparsity levels according to the importance of corresponding timestep regions, allowing for aggressive sparsity during non-critical regions while conserving computational resources for crucial timestep regions.  
Building on this, we develop an efficient DM training technique that derives timestep-aware EB tickets, trains them in parallel, and combines them during inference for image generation.  Extensive experiments validate the existence of both traditional and timestep-aware EB tickets, as well as the effectiveness of our proposed EB-Diff-Train method. This approach can significantly reduce training time both spatially and temporally—achieving 2.9$\times$$\sim$5.8$\times$ speedups over training unpruned dense models, and up to 10.3$\times$ faster training compared to standard train-prune-finetune pipelines—without compromising generative quality. 
 Our code is available at \href{https://github.com/GATECH-EIC/Early-Bird-Diffusion}{https://github.com/GATECH-EIC/Early-Bird-Diffusion}.
\end{abstract}
\begin{figure}[ht!]
    \centering
    \includegraphics[height=3in, width=\columnwidth, keepaspectratio]{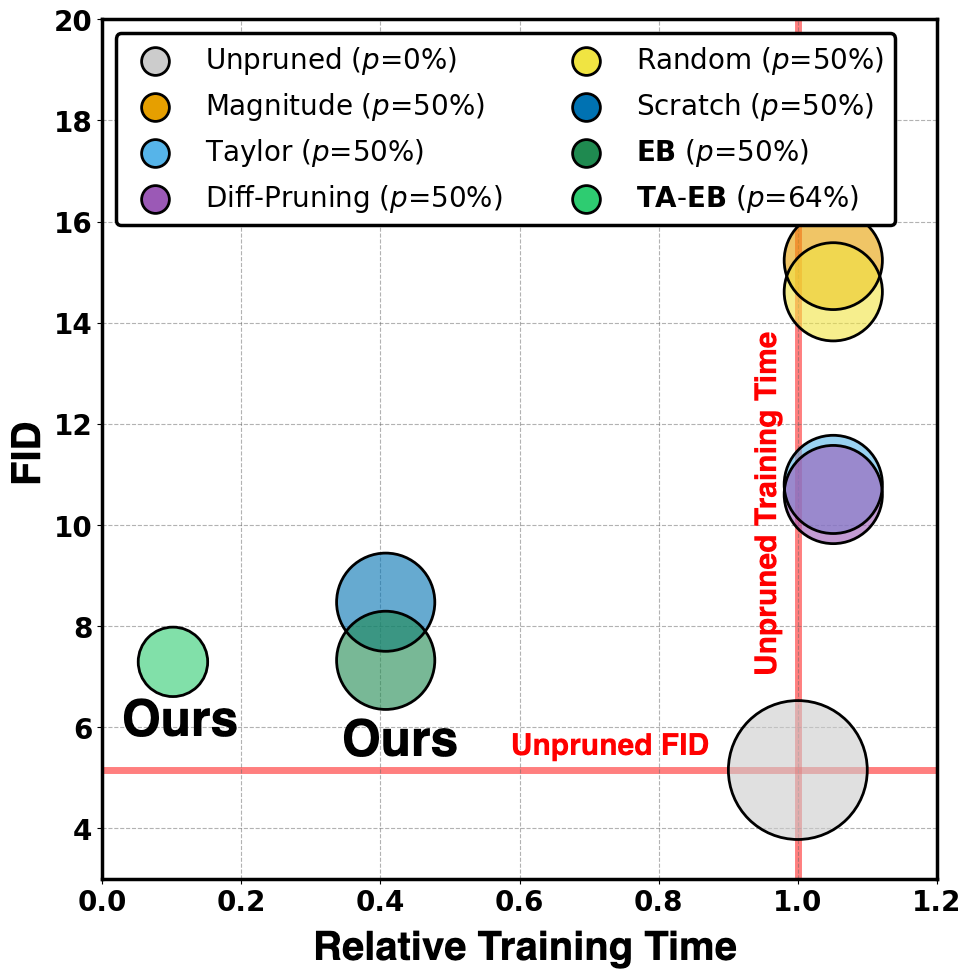}
    \vspace{-1em}
\caption{
FID (lower values indicate higher generation quality) vs. relative training time (lower values indicate greater runtime efficiency) under different pruning rates (``$p$'') for the CIFAR-10 \cite{cifar10} dataset and the DDPM \cite{ho2020denoising} model, comparing our methods against the standard train-prune-finetune method paired with random \cite{frankle2018the}, magnitude \cite{hanMagnitudePrune}, Taylor \cite{Molchanov_2019_CVPR}, and Diff-Pruning \cite{diffPruning} pruning methods. ``Scratch" indicates a model pruned then retrained from random initialization, ``Unpruned" is the model without any pruning, ``EB" (as detailed in Sec. 4.1) uses a single EB ticket across all timesteps, and ``TA-EB" (as detailed in Sec. 4.2) employs three timestep-aware EB tickets for specific regions. Smaller circles indicate higher pruning rates; relative training time represents the ratio to the unpruned model's training time. Generation quality is measured by the FID \cite{fid_score} score.}
\vspace{-1em}
    \label{fig:fid_vs_flops}
\end{figure}

\vspace{-1.5em}
\section{Introduction}
Diffusion Models (DMs) \cite{ho2020denoising, deepUnsupervisedLearningNonequilibrium, song2021scorebased} have become powerful generative tools, transforming noise into structured samples and achieving state-of-the-art (SOTA) generation quality \cite{diffModelsBeatGANs}. Training these models, however, is prohibitively expensive; popular models like RAPHAEL \cite{xue2024raphaeltexttoimagegenerationlarge} and DALL·E 2 \cite{openai2022dalle2} can require tens of thousands of A100 GPU days, leading to significant financial and time costs. This challenge has spurred the development of various efficient diffusion training techniques such as assigning different weights to each timestep \cite{hang2024efficientdiffusiontrainingminsnr,wang2024closer,Choi_2022_CVPR}, introducing new model architectures \cite{rombach2021highresolution, lee2023multiarchitecturemultiexpertdiffusionmodels, balaji2023ediffitexttoimagediffusionmodels}, and employing pruning strategies \cite{diffPruning,zhang2024laptopdifflayerpruningnormalized,kim2023bksdmlightweightfastcheap}. Although these strategies alleviate some computational burdens, they typically still necessitate fully training an unpruned model as an initial step, incurring substantial initial training costs.

In parallel, the ``Lottery Ticket Hypothesis" (LTH) \cite{frankle2018the} suggests the existence of ``winning tickets", sparse subnetworks that, when fully trained, can perform as well or better than the original dense network. Extending this concept, ``Early-Bird" (EB) tickets \cite{earlyBird} have been developed to identify high-performing subnetworks that emerge early in training, reducing the need to fully train a dense model to convergence. This concept has shown the potential to significantly lower training costs across various machine learning models \cite{earlyBird, gan_eb, you2021earlybirdgcnsgraphnetworkcooptimization,chen2020earlybert}.

Building on the insights from DM pruning and EB training, we explore the possibility of exceeding the efficiency vs. quality trade-offs of the current train-prune-finetune pipeline \cite{diffPruning} for DMs without fully training a DM model. In addition, previous studies \cite{balaji2023ediffitexttoimagediffusionmodels, lee2023multiarchitecturemultiexpertdiffusionmodels, wang2024closer} have shown that different timestep regions contribute variably to the denoising process, with some regions proving easier to learn than others. By harnessing these unique properties of DMs, we aim to significantly reduce training time without compromising on generation quality. We pose the following key research questions: \textbf{Q1:} \textit{Do EB tickets exist in DMs?} \textbf{Q2:} \textit{Can we identify and leverage diffusion-dedicated timestep-aware EB tickets for efficient training?}
In summary, our contributions are as follows: 
\begin{enumerate} 

\item \textit{Identification of traditional EB Tickets in DMs:} We empirically validate the consistent presence of EB tickets in DMs, demonstrating that high-performing sparse subnetworks emerge early in the training process. This investigation, to the best of our knowledge, is the first to explore the EB phenomenon within the context of DMs and validate its enhanced training efficiency for DMs. \vspace{0.1em} 

\item \textit{Investigation of diffusion-dedicated Timestep-Aware EB (TA-EB) Tickets:} We introduce and characterize distinct EB tickets tailored for different timestep regions. These tickets adapt their sparsity levels according to the importance of corresponding timestep regions, offering the potential for aggressive sparsity during non-critical regions while less pruning for critical ones. 

\item \textit{Development of EB-Diff-Train:} Building on the concept of TA-EB tickets, we develop the EB-Diff-Train method to train these TA-EB tickets in parallel and then ensemble them during inference for image generation. This method reduces the overall training time through temporal parallelism and spatially tailored subnetworks, leveraging the unique advantages of each TA-EB subnetwork 
to form an efficient training strategy that maintains high-quality generations, as demonstrated in Figure \ref{fig:fid_vs_flops}. 

\end{enumerate}

We conduct extensive experiments and ablation studies to validate the existence of both traditional EB tickets and diffusion-dedicated TA-EB tickets in DMs, as well as the effectiveness of the proposed EB-Diff-Train using both EB and TA-EB tickets. We believe this work contributes to a deeper understanding of the training process for DMs.
\vspace{-0.5em}
\section{Related Works}
\textbf{Diffusion Models.}  
DMs are a class of generative models that transform random noise into structured data by reversing a forward diffusion process, which gradually corrupts data over a series of timesteps \cite{ho2020denoising, song2021scorebased, deepUnsupervisedLearningNonequilibrium}. The primary goal of these models is to learn the reverse process by iteratively denoising samples to recover the original data distribution. This has been formalized as a denoising score matching problem \cite{hyvarianScoreMatching}, where models learn the conditional distributions required at each step of the reverse process. Recent works \cite{ho2020denoising, song2019generative, song2019sliced, song2021scorebased} have explored this in depth. As the noising process is controlled by the timestep domain, some timesteps are particularly informative for maintaining content and structure, influencing the design and optimization of DMs \cite{balaji2023ediffitexttoimagediffusionmodels, Zhang_2024_CVPR, Choi_2022_CVPR, NEURIPS2023_56a7b9a0, go2023towards}.

\textbf{Efficient Diffusion Model Training.} Current strategies to streamline diffusion model training can be grouped into three categories: sampling techniques, training techniques, and architectural enhancements. Sampling techniques aim to enhance efficiency by reducing the number of denoising steps or by optimizing timestep selection. For instance, methods utilizing Stochastic Differential Equation (SDE) and Ordinary Differential Equation (ODE) solvers \cite{song2021scorebased, lu2022dpmsolverfastodesolver} enhance sample generation through optimized integration steps. Furthermore, timestep weighting methods adjust loss weights to refine sampling efficiency, focusing on metrics such as the signal-to-noise ratio \cite{hang2024efficientdiffusiontrainingminsnr} or perceptual detail \cite{Choi_2022_CVPR}. On the training front, techniques like partial image training \cite{ding2023patcheddenoisingdiffusionmodels, wang2023patchdiffusionfasterdataefficient} and wavelet-based diffusion \cite{phung2023waveletdiffusionmodelsfast} have been developed to reduce computational costs. Architectural enhancements also play a crucial role, with innovations such as performing the diffusion process in latent space to drastically improve training feasibility \cite{rombach2021highresolution}. Additionally, modifications to the standard diffusion model architecture, including the incorporation of decoders or expert networks \cite{Zhang_2024_CVPR, go2023towards}, target various noise ranges to optimize performance. In contrast to these methods, our EB-Diff-Train approaches enable significant reductions in training demands by identifying effective subnetworks early in the training process, offering an orthogonal method to enhance the training efficiency of diffusion models.

\textbf{Diffusion Model Pruning.} Pruning strategies for diffusion models have recently gained attention. \cite{diffPruning} introduced a pruning method that utilizes a Taylor expansion of pruned timesteps to aggregate informative gradients, thereby identifying crucial connections across different timestep regions. \cite{zhang2024laptopdifflayerpruningnormalized} proposed a one-shot criterion for layer pruning in DMs, which includes distillation-based retraining to recover performance after pruning. Meanwhile, \cite{kim2023bksdmlightweightfastcheap} focused on block pruning coupled with feature distillation to decrease the number of parameters while maintaining comparable performance. Our EB-Diff-Train approaches can leverage these pruning methods to identify efficient, high-performing subnetworks.

\begin{figure}[!t]
    \centering
    \includegraphics[width=\columnwidth, keepaspectratio]
{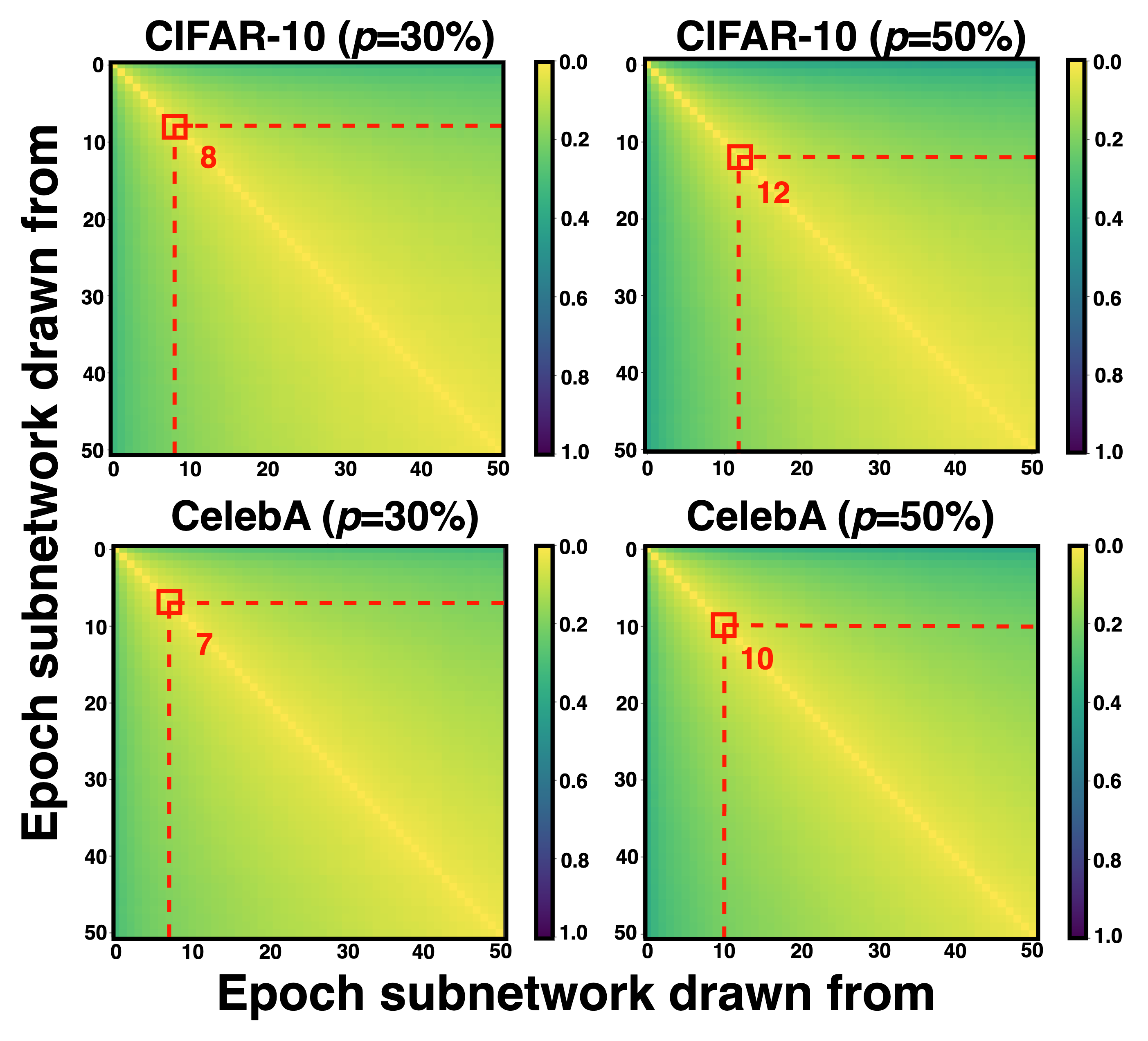}
    \caption{Visualization of pairwise hamming distance matrices for both the CIFAR-10 \cite{cifar10} and CelebA \cite{celeba} datasets, when using structural magnitude pruning at pruning rates of 30\% and 50\%. EB tickets (marked by red boxes) are consistently found during the early stages of training.}
    \label{fig:pairwise_cifar10_celeba_trad}
    \vspace{-1.5em}
\end{figure}

\textbf{Lottery Ticket Hypothesis and Early-Bird Tickets.}
The \textit{Lottery Ticket Hypothesis} \cite{frankle2018the} posits that within large, densely initialized neural networks lie smaller subnetworks—referred to as ``winning tickets''—that are capable of achieving comparable or superior performance when trained independently. However, identifying these winning tickets traditionally requires fully training the dense model first, thus limiting the efficiency gains. Addressing this limitation, \cite{earlyBird} introduced the concept of \textit{Early-Bird tickets} (``EB tickets''), which identifies high-performing subnetworks early in the training process. This is achieved by training a dense model for a few epochs and monitoring its pruned masks, stored in a first-in first-out (FIFO) queue. When the Hamming distances of these masks stabilize below a certain threshold $\eta$, it indicates a stable pruned subnetwork (i.e., an ``EB Ticket''), which can then be extracted and fully trained in place of the dense network. This technique has been successfully applied in various architectures such as CNNs \cite{earlyBird}, VAEs and GANs \cite{gan_eb}, and graph neural networks \cite{you2021earlybirdgcnsgraphnetworkcooptimization}, allowing for sparse networks that perform comparably to or even better than their dense counterparts.

\section{Preliminaries of Diffusion Models}

\textbf{Diffusion Models.} DMs have emerged as a powerful class of generative models that synthesize data by learning to reverse a stochastic degradation process \cite{ho2020denoising,deepUnsupervisedLearningNonequilibrium,song2019generative}. The core mechanism involves two phases: a forward process that systematically injects noise into data samples, and a learned reverse process that reconstructs the original distribution through iterative denoising.

In the forward process, given an initial sample $x_{0}$ from data distribution $q$, the data undergoes a progressive corruption over $T$ timesteps according to a predefined variance schedule $\{\alpha_{t},\sigma_t \}_{t=1}^{T}$. At each timestep $t$, the corrupted sample follows a Gaussian distribution:
\vspace{-0.5em}
\[
q(x_t \mid x_0) = \mathcal{N}(\alpha_t x_0, \sigma_t^{2} \mathbf{I}) \tag{1},
\vspace{-0.5em}
\]
where the sample $x_t$ is obtained through a reparameterization using Gaussian noise $\epsilon \sim \mathcal{N}(0,\mathbf{I})$:
\vspace{-0.5em}
\[
x_t = \alpha_t x_0 + \sigma_t \epsilon \tag{2}.
\vspace{-0.5em}
\]

This process continues until timestep $T$, where $x_T$ converges to a Gaussian distribution: $x_T \sim \mathcal{N}(0,\mathbf{I})$. The generation process learns to invert this corruption by training a noise prediction network $\epsilon_\psi (x_t,t)$. Starting from pure noise $x_T \sim \mathcal{N}(0,\mathbf{I})$, the model iteratively denoises the sample through $T$ reverse steps. The parameters $\psi$ are optimized by minimizing the objective:
\vspace{-0.5em}
\[
    \mathcal{L}(\psi) = \mathbb{E}_{t,\epsilon \sim\mathcal{N}(0,\mathbf{I})} \left[|\epsilon_\psi(x_{t},t) - \epsilon|_2^{2}\right].\tag{3}
\]
\section{Our Findings and Proposed Techniques}
\label{sec:method}

In this section, we explore the existence of both traditional and diffusion-dedicated EB tickets. DMs are distinct from other generative models like GANs \cite{goodfellow2014generativeadversarialnetworks} and VAEs \cite{kingma2022autoencodingvariationalbayes} due to their reliance on iterative denoising. Recent studies \cite{balaji2023ediffitexttoimagediffusionmodels,Zhang_2024_CVPR,go2023towards,lee2023multiarchitecturemultiexpertdiffusionmodels} have utilized this characteristic to efficiently train separate models for distinct timestep regions. Inspired by this approach, we initially investigate the presence of traditional EB tickets in DMs, which identify a single effective subnetwork across all timesteps, addressing \textbf{Q1}. Subsequently, we develop and identify diffusion-dedicated TA-EB tickets, which create distinct EB subnetworks for different timestep regions, addressing \textbf{Q2}. Leveraging these findings, we devise an efficient training technique that can reduce training costs both temporally and spatially throughout the training trajectory without negatively impacting the generation quality.

\subsection{Q1: Do EB Tickets Exist in Diffusion Models?}
\label{sec:Q1}
In this subsection, we seek to explore and validate the existence of traditional EB tickets in DMs. 

\begin{table}[t]
\centering
\caption{EB Tickets with 30\% and 50\% pruning rates on CIFAR-10 with 32 $\times$ 32 resolution using the DDPM \cite{ho2020denoising}. The best FID score is highlighted in \textbf{bold}. ``Early-Bird'' is the epoch in which the EB ticket is identified.}
\vspace{-0.5em}
\scriptsize 
\setlength{\tabcolsep}{2pt} 
\resizebox{\linewidth}{!}{ 
\begin{tabular}{@{}lcccccc@{}}
\toprule
\multirow{2}[2]{*}{\tabincell{l}{\textbf{Pruning}\\\textbf{Metric}}} & \multicolumn{5}{c}{\textbf{DDPM @ CIFAR-10 (32$\times$32)}} \\
\cmidrule(lr){2-6}
 & \textbf{\#Params} $\downarrow$ & \textbf{MACs} $\downarrow$ & \textbf{FID} $\downarrow$ & \textbf{Early-Bird} & \textbf{Iters} \\
\midrule
\multicolumn{6}{c}{$\textit{\textbf{30\%} Pruning Rate}$} \\
\cmidrule(lr){2-4}
\multirow{4}{*}{\shortstack[l]{Magnitude \\
Taylor \\
Diff-Pruning \\
\textbf{Iter-wise Magnitude}}} 
& \multirow{4}{*}{19.8M} & \multirow{4}{*}{3.4G} 
& 5.71 & 4 & 900K\\
 & & & 5.68 & 25 & 900K  \\
 & & & 6.10 & 26 & 900K\\
 & & & \textbf{5.33} & \textbf{877} & \textbf{900K} \\
\midrule
\multicolumn{6}{c}{$\textit{\textbf{50\%} Pruning Rate}$} \\
\cmidrule(lr){2-4}
\multirow{4}{*}{\shortstack[l]{\textbf{Magnitude} \\
Taylor \\
Diff-Pruning \\
Iter-wise Magnitude}} 
& \multirow{4}{*}{9.0M} & \multirow{4}{*}{1.5G} 
& \textbf{7.32} & \textbf{4} & \textbf{900K} \\
 & & & 7.69 & 25 & 900K \\
 & & & 8.20 & 25 & 900K \\
 & & & 7.46 & 1144 & 900K \\
\bottomrule
\vspace{-2.5em}
\end{tabular}
}
\label{tab:findings_cifar10_trad}
\vspace{-1em}
\end{table}

\textbf{Settings.} In this study, we employ the structural pruning method from \cite{diffPruning} to prune Denoising Diffusion Probabilistic Models (DDPMs) \cite{ho2020denoising}, using representational datasets including CIFAR-10 \cite{cifar10} and CelebA \cite{celeba}. We explore various pruning strategies including magnitude \cite{hanMagnitudePrune}, Taylor \cite{Molchanov_2019_CVPR}, and Diff-Pruning \cite{diffPruning}. For Diff-Pruning, we set a threshold of $\mathcal{T}=0.05$ to balance performance and model simplicity. We evaluate pruning at 30\% and 50\% pruning rates for CIFAR-10, and at 50\% for CelebA, assessing performance via the Fréchet Inception Distance (FID) \cite{fid_score}. All hyperparameters are aligned with those specified in \cite{diffPruning}, and image generation is performed using 100 DDIM \cite{song2020denoising} steps. Following the procedure outlined in \cite{earlyBird} to find EB tickets, we save epoch-wise pruned masks during training and monitor their convergence. Upon convergence, we extract and fully train the resulting subnetwork instead of the original dense model. Unless specified otherwise, we use a convergence threshold of $\eta=0.1$ and a FIFO queue length of 5 for storing and evaluating pruned masks.

\textbf{Observations.} Figure \ref{fig:pairwise_cifar10_celeba_trad} illustrates the pairwise mask distances for CIFAR-10 \cite{cifar10} and CelebA \cite{celeba} at pruning rates of 30\% and 50\%, employing magnitude-based pruning. Here, the $(i,j)$-th element represents the distances between the pruned subnetworks sampled at the $i$-th and $j$-th epochs. Lighter colors indicate lower inter-mask Hamming distances, while darker colors denote higher distances. The epoch in which the EB ticket was identified is highlighted in red font. Notably, EB tickets are consistently identified in the early stages of training, demonstrating that such subnetworks can be found with minimal extra cost—approximately 0.5\%$\sim$2.5\% of the total training iterations. Tables \ref{tab:findings_cifar10_trad} and \ref{tab:findings_celeba_trad} present the results for our EB tickets across different pruning rates and methods. To compare against the standard epoch-wise EB training done in \cite{earlyBird}, we also explore iteration-wise EB training in our CIFAR-10 examples, where we monitor the pruned masks every iteration instead of every epoch. According to Tables \ref{tab:findings_cifar10_trad} and \ref{tab:findings_celeba_trad}, although the performances of the EB tickets found are generally comparable, magnitude pruning consistently identifies the earliest EB tickets. Given that other pruning methods require costly expansions of the loss over timesteps and our objective is to identify high-performing subnetworks with the least computational expense, we will restrict our use to magnitude pruning in subsequent experiments involving larger datasets.

\textbf{Our Answer to Q1.} Through this series of experiments, we have consistently identified and observed EB tickets across various datasets, pruning rates, and pruning strategies. Importantly, we see that these tickets emerge early in training and have minimal overhead cost to find.
Similar experiments for other diffusion models and datasets can be found in the supplementary material.

\begin{figure*}[t]
    \centering
    \includegraphics[width=1.0\linewidth, keepaspectratio]{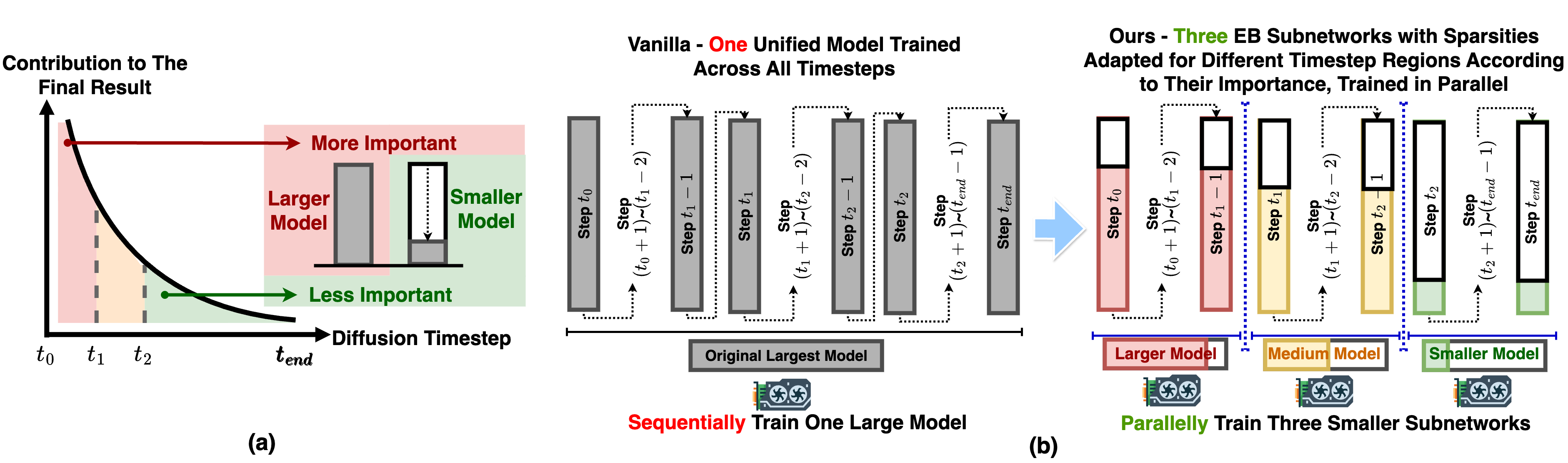}
    \vspace{-2em}

\caption{(a) The varying importance of timestep regions throughout the diffusion training trajectory \cite{wang2024closer}, which motivates our investigation into TA-EB tickets. (b) A comparison between vanilla training of DMs and our proposed TA-EB training, which offers the dual advantages of reducing model size through dedicated EB subnetworks and enhancing parallelism, thereby yielding savings both spatially and temporally.}
\vspace{-1.5em}
    \label{fig:comparison_methods}
\end{figure*}

\begin{table}
\centering
\caption{EB Tickets with 50\% pruning rate on CelebA with 64 $\times$ 64 resolution using the DDPM \cite{ho2020denoising}. The best FID score is highlighted in \textbf{bold}. ``Early-Bird'' is the epoch in which the EB ticket is identified.}
\vspace{-0.5em}
\scriptsize 
\setlength{\tabcolsep}{4pt} 
\renewcommand{\arraystretch}{1.2} 
\resizebox{\linewidth}{!}{ 
\begin{tabular}{@{}lcccccc@{}}
\toprule
\multirow{2}[2]{*}{\tabincell{l}{\textbf{Pruning}\\\textbf{Metric}}} & \multicolumn{5}{c}{\textbf{DDPM @ CelebaA (64$\times$64)}}
\\
\cmidrule(lr){2-6}
 & \textbf{\# Params} $\downarrow$ & \textbf{MACs} $\downarrow$ & \textbf{FID} $\downarrow$ & \textbf{Early-Bird} & \textbf{Iters} \\
\midrule
\multicolumn{6}{c}{$\textit{\textbf{50\%} Pruning Rate}$} \\
\cmidrule(lr){2-4}
\multirow{6}{*}{\shortstack[l]{Magnitude \\ Magnitude \\ Taylor \\ \textbf{Taylor} \\ Diff-Pruning\\  Diff-Pruning}} 
& \multirow{6}{*}{19.7M} & \multirow{6}{*}{6.0G} & 5.61 & 15 & 500K\\
 & & & 5.53 & 15 & 600K\\
 & & & 5.50 & 24 & 500K\\
 & & & \textbf{5.42} & \textbf{24} & \textbf{600K}\\
 & & & 5.66 & 25 & 500K\\
 & & & 5.58 & 25 & 600K\\
\bottomrule
\vspace{-2.5em}
\end{tabular}
}
\label{tab:findings_celeba_trad}
\vspace{-1em}
\end{table}

\vspace{-0.2em}
\subsection{Q2: Can We Identify and Leverage Diffusion-Dedicated EB Tickets for Efficient Training?}
\label{sec:Q2}

In this subsection, we first investigate the existence of diffusion-dedicated TA-EB tickets by examining the unique characteristics of DMs and their training processes. We then develop and apply an efficient training technique for DMs leveraging our identified diffusion-dedicated TA-EB tickets.

\textbf{Characteristics of DM Training and Our Hypothesis.} The denoising process in DMs is governed by the number of timesteps, with later timesteps characterized by lower signal-to-noise ratios and consequently less informational content \cite{hang2024efficientdiffusiontrainingminsnr, balaji2023ediffitexttoimagediffusionmodels}. In response to this, several studies have introduced architectural adaptations, utilizing different models or model components across specific timestep regions \cite{lee2023multiarchitecturemultiexpertdiffusionmodels, balaji2023ediffitexttoimagediffusionmodels, oms_dpm}. Inspired by these adaptations, we hypothesize the feasibility of extracting timestep-aware EB tickets—EB subnetworks specifically tailored to different timestep regions. These dedicated subnetworks can be trained in parallel, thus optimizing the training process by exploiting the unique characteristics of each timestep region. This approach aims to significantly reduce training costs both temporally (i.e., timestep-level) and spatially (i.e., model-level), offering a more efficient way to train SOTA DMs.

\begin{figure}[b!]
    \centering
    \vspace{-2em}
    \includegraphics[width=\columnwidth, keepaspectratio]{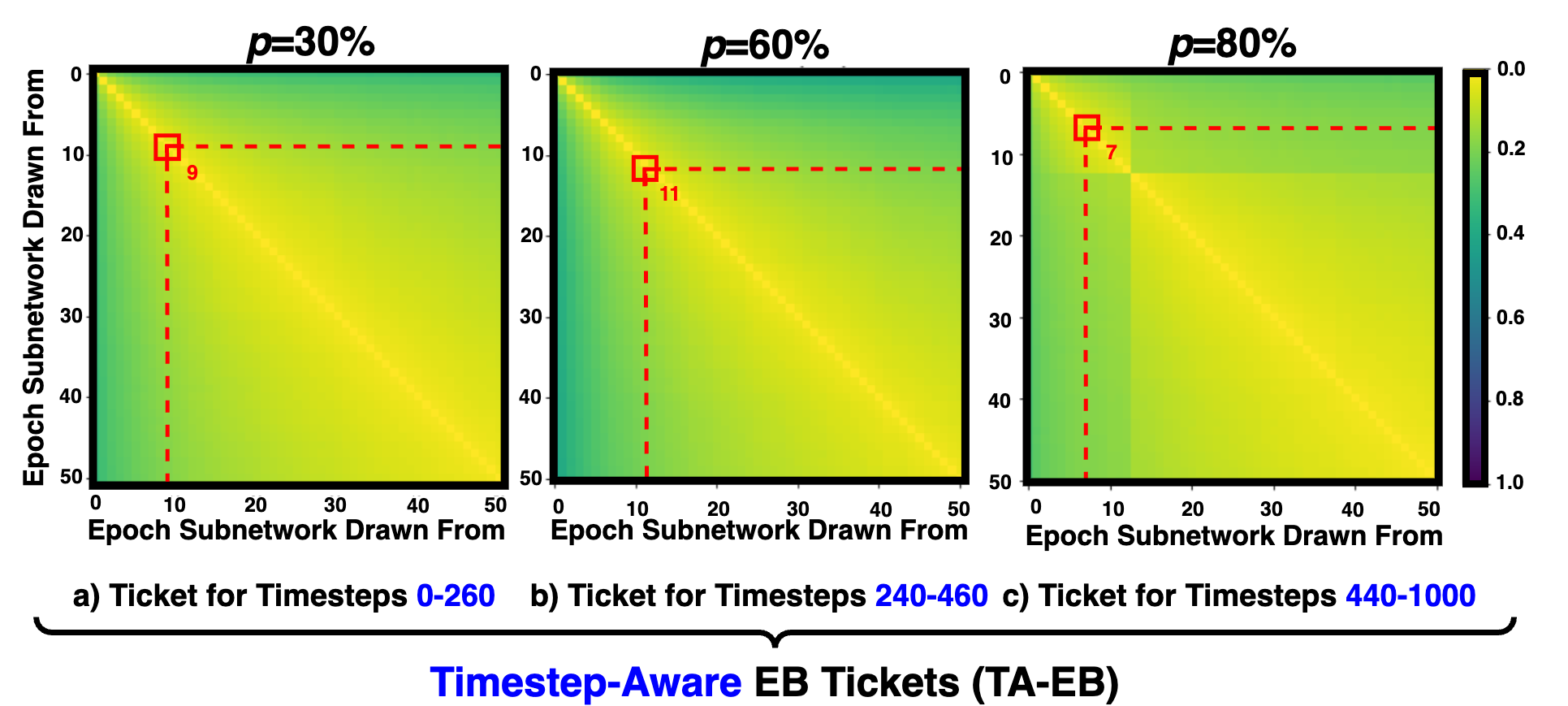}
\vspace{-2em}
    \caption{Visualization of pairwise Hamming distance matrices for the CIFAR-10~\cite{cifar10} dataset under structural magnitude pruning at pruning rates of 30\%, 60\%, and 80\% across timestep regions 0-260, 240-460, and 440-1000, respectively. EB tickets are consistently observed during the early stages of each timestep region.}
    \label{fig:timestep_aware_ticket_heatmaps}
    \vspace{-1em}
\end{figure}

\textbf{Settings.} To effectively draw TA-EB tickets, a crucial initial step is partitioning the training trajectory timesteps into appropriate regions. We leverage insights from \cite{wang2024closer}, which categorizes the timesteps into three regions: acceleration, deceleration, and convergence. The insight from \cite{wang2024closer} is that the convergence region, typically oversampled, contains less dynamic learning challenges compared to the acceleration and deceleration regions, as shown in Figure \ref{fig:comparison_methods} (a); To optimize the training process, sampling from the convergence region is minimized, while emphasis is increased on the timesteps within the acceleration and deceleration regions. Additionally, to establish smooth transitions between these regions and mitigate overfitting issues identified in \cite{hang2024efficientdiffusiontrainingminsnr}, we introduce a slight overlap of 2\% of timesteps at each boundary. Empirical results, as illustrated in Table \ref{tab:timestep_regions}, show that this small overlap improves the FID scores.
In this experiment, we utilize the CIFAR-10 \cite{cifar10} and CelebA \cite{celeba} datasets with DDPMs \cite{ho2020denoising} as the model framework. The resulting pruning rates of EB tickets for the three designated regions are as follows: low pruning rate of 30\% in Region 1 \((0 \leq t \leq 260)\), medium pruning rate of 60\% in Region 2 \((240 \leq t \leq 460)\), and high pruning rate of 80\% in Region 3 \((440 \leq t \leq 1000)\).

\textbf{Observations.} Figure~\ref{fig:timestep_aware_ticket_heatmaps} illustrates the pairwise mask distances for CIFAR-10 \cite{cifar10} across designated timestep regions, using magnitude-based pruning at rates of 30\%, 60\%, and 80\%, respectively, adapting to their importance to the final result (see Figure \ref{fig:comparison_methods} (a)). Similar to prior demonstrations (see Figure \ref{fig:pairwise_cifar10_celeba_trad}), the $(i,j)$-th element of each matrix captures the Hamming distances between subnetworks pruned at epochs $i$ and $j$, with lighter colors denoting lower distances, indicating more similar subnetwork structures.
Interestingly, these visualizations validate our hypothesis regarding the existence of timestep-aware EB (TA-EB) tickets—subnetworks that are specifically tailored to operate efficiently within designated timestep regions of the training process, i.e., validating the presence of EB tickets dedicated to timestep training of DMs. Consistent with our expectations and similar to traditional EB findings, TA-EB tickets are observed predominantly in the early stages of each timestep region. Similar experiments for other diffusion models/datasets can be found in the supplements.

\begin{table}[t!]
\centering
\caption{Comparison between overlapping and non-overlapping timestep regions in terms of FID. The best FID is \textbf{bolded}.}
\vspace{-0.5em}
\resizebox{\columnwidth}{!}{%
\begin{tabular}{c c c c}
    \toprule
    \textbf{Region 1 Bounds} & \textbf{Region 2 Bounds} & \textbf{Region 3 Bounds} & \textbf{FID} $\downarrow$ \\ 
    \midrule
    0-240 & 240-440 & 440-1000 & 8.10 \\
    \textbf{0-260} & \textbf{240-460} & \textbf{440-1000} & \textbf{7.69} \\
    \bottomrule
    \vspace{-3em}
\end{tabular}%
}
\label{tab:timestep_regions}
\end{table}

\textbf{Proposed EB-Diff-Train Method.} The identification of TA-EB tickets facilitates the extraction of effective subnetworks, with sparsity levels tailored to the significance of corresponding timestep regions during the training trajectory. Instead of relying on a single model or subnetwork configuration throughout the entire training process, these timestep-aware EB tickets offer the flexibility to customize subnetwork sparsities for each timestep region, thereby optimizing computational resources and enhancing training parallelism. We hypothesize that training these region-specific TA-EB tickets in parallel, and subsequently integrating them at the end of their respective training periods, will not compromise the generative quality of the derived model. This hypothesis is based on the premise that EB subnetworks trained for different timestep region are likely to learn complementary features and capabilities, thereby contributing to the robustness and generalization capabilities of the final model. Specifically, our EB-Diff-Train method consistently achieves lower FID scores (the lower, the better) than both the vanilla EB training and scratch training methods in both datasets, utilizing timestep-aware EB tickets visualized in Figure \ref{fig:timestep_aware_ticket_heatmaps}.

We refer to this strategy of drawing region-specific EB tickets, training them concurrently, and then combining them through an ensemble as EB-Diff-Train. Figure \ref{tab:timestep_regions} presents an illustrative comparison between commonly used scratch DM training and our EB-Diff-Train method, which promises training savings both spatially and temporally.
Our EB-Diff-Train approach aims to reduce overall training time through temporal parallelism and strategically tailored spatial EB subnetworks, leveraging the distinct advantages of each subnetwork to produce a more capable model from a more efficient training pipeline.

\textbf{Our Answer to Q2.} Through our investigations and experiments above and in Section \ref{sec:exp}, we conclude that diffusion-dedicated TA-EB tickets indeed exist. These tickets are drawn from timestep region-specific insights, allowing for the customization of subnetwork sparsity ratios according to the significance of corresponding timestep regions. Furthermore, the existence of these TA-EB tickets presents opportunities to develop more efficient DM training techniques. Our EB-Diff-Train method exemplifies one such implementation.
\begin{table}[t!]
\centering
\caption{Results for CIFAR-10 32$\times$32 using the DDPM \cite{ho2020denoising} model with 100 DDIM steps. The best FID is \textbf{bolded}. The number of iterations for the \textbf{TA-EB} methods are recorded as the total number of iterations for the subnetwork of longest training time.}
\vspace{-0.5em}
\scriptsize 
\setlength{\tabcolsep}{1.5pt} 
\resizebox{\columnwidth}{!}{ 
\begin{tabular}{@{}lcccccc@{}}
\toprule
\multirow{2}[2]{*}{\tabincell{l}{\textbf{Pruning}\\\textbf{Metric}}} & \multicolumn{5}{c}{\textbf{DDPM @ CIFAR-10 (32$\times$32)}} \\ 
\cmidrule(lr){2-6}
& \textbf{\#Params} $\downarrow$ & \textbf{MACs} $\downarrow$ & \textbf{FID} $\downarrow$ & \textbf{Iters} & \textbf{Speed-Up} $\uparrow$\\
\midrule
Unpruned & 35.7M & 6.1G & 5.15 & 800K & 1.00$\times$\\
\midrule
\multicolumn{6}{c}{\textbf{Baselines} w/ 30\% Pruning Rate} \\
\cmidrule(lr){2-5}
\multirow{5}{*}{\shortstack[l]{Scratch \\ Random \\ Magnitude \\ Taylor \\ Diff-Pruning}} 
& \multirow{5}{*}{19.8M} & \multirow{5}{*}{3.4G} 
& 5.45 & 900K & 0.56$\times$\\
 & & & 6.61 &  100K & 0.92$\times$\\
 & & & 6.72 &  100K & 0.92$\times$\\
 & & & 6.36 & 100K & 0.92$\times$\\
 & & & 6.32 &  100K & 0.92$\times$\\
\midrule
\multicolumn{6}{c}{\textbf{Baselines} w/ 50\% Pruning Rate} \\
\cmidrule(lr){2-5}
\multirow{5}{*}{\shortstack[l]{Scratch \\ Random \\ Magnitude \\ Taylor \\ Diff-Pruning}} 
& \multirow{5}{*}{9.0M} & \multirow{5}{*}{3.4G} 
& 8.47 & 900K & 0.71$\times$\\
 & & & 14.61 & 100K & 0.95$\times$\\
 & & & 15.23 &  100K & 0.95$\times$\\
 & & & 10.75 &  100K & 0.95$\times$\\
 & & & 10.68 &  100K & 0.95$\times$\\
\midrule
\multicolumn{6}{c}{\textbf{EB-Diff-Train (EB)} w/ 30\% Pruning Rate} \\
\cmidrule(lr){2-5}
\multirow{4}{*}{\shortstack[l]{Magnitude \\ Taylor \\ Diff-Pruning \\ Iter-wise Mag.}} 
& \multirow{4}{*}{19.8M} & \multirow{4}{*}{3.4G} 
& 5.71 & 900K  & 1.28$\times$\\
 & & & 5.68  & 900K & 1.28$\times$\\
 & & & 6.10  & 900K & 1.28$\times$\\
 & & & \textbf{5.33}  & 900K & \textbf{1.28$\times$}\\
\midrule
\multicolumn{6}{c}{\textbf{EB-Diff-Train (EB)} w/ 50\% Pruning Rate} \\
\cmidrule(lr){2-5}
\multirow{4}{*}{\shortstack[l]{Magnitude \\ Taylor \\ Diff-Pruning \\ Iter-wise Mag.}} 
& \multirow{4}{*}{9.0M} & \multirow{4}{*}{3.4G} 
& \textbf{7.32} & 900K & \textbf{2.45$\times$}\\
 & & & 7.69  & 900K & 2.45$\times$\\
 & & & 8.20  & 900K & 2.45$\times$\\
 & & & 7.46 & 900K & 2.45$\times$\\
\midrule
\multicolumn{6}{c}{\textbf{EB-Diff-Train (TA-EB)} w/ Avg. 64\% Pruning Rate} \\
\cmidrule(lr){2-5}
 Magnitude & 7.2M & 1.3G & \textbf{7.29} & 200K & \textbf{5.78}$\times$\\
\bottomrule

\end{tabular}
}

\label{tab:cifar10_results_trad}
\end{table}

\vspace{-0.7em}
\section{Experiments}
\label{sec:exp}
\vspace{-0.5em}

In this section, we empirically validate the effectiveness of our proposed EB-Diff-Train by comparing it with the SOTA training baseline across five datasets and two diffusion models. To differentiate between the two types of EB tickets, we refer to those drawn according to Sec.~\ref{sec:Q1} as ``EB-Diff-Train (EB)'' and those following Sec.~\ref{sec:Q2} as ``EB-Diff-Train (TA-EB)'', which denote tickets that apply varying pruning rates across different timestep regions.
\vspace{-0.5em}
\subsection{Experiment Settings}
\label{sec:exp_settings}
\textbf{Models and Datasets.} We evaluate our EB and TA-EB tickets on five representative datasets: CIFAR-10~\cite{cifar10}, CelebA~\cite{celeba}, LSUN Bedroom~\cite{lsun}, LSUN Church~\cite{lsun}, and ImageNet-1K~\cite{deng2009imagenet}. For these evaluations, we use two popular diffusion models: DDPM~\cite{ho2020denoising} for CIFAR-10, CelebA, and the LSUN datasets, and LDM~\cite{rombach2021highresolution} for ImageNet-1K.

\textbf{Training and Sampling Settings.} 
We use the same hyperparameters and model setup as Diff-Pruning~\cite{diffPruning}, with the only difference being a reduction in batch size from 64 to 8 for LSUN Church, LSUN Bedroom, and ImageNet-1K to enable training on a single A100 GPU.

\textbf{Baselines and Evaluation Metrics.} 
\underline{\textit{Baselines.}} We compare the proposed EB-Diffusion against SOTA baselines, including Diff-Pruning~\cite{diffPruning}, Taylor pruning~\cite{Molchanov_2019_CVPR}, magnitude and random pruning~\cite{earlyBird,frankle2018the}. 
Given that training pruned networks from scratch is proven to be highly competitive~\cite{diffPruning}, we also include a pruned network trained from scratch in our comparisons, referred to as ``Scratch''. In this approach, we first prune a fully trained network to determine its connectivity, then reinitialize it with random weights for training.
\underline{\textit{Evaluation Metrics.}} We compare image generation quality using the FID metric \cite{fid_score}, efficiency as measured by the number of parameters (\#Params) and Multiply-Add Accumulation operations (MACs). We also record the training speedup, which is the ratio of the unpruned model's training time to each pruned model's training time, as measured on the same A100 GPU. For our TA-EB tickets, we measure the \#Params and MACs as the weighted average distributed across timesteps. 

\subsection{Our EB-Diff-Train over SOTA Baselines} 

\textbf{Results for CIFAR-10.} To assess the effectiveness of our proposed EB-Diff-Train, we apply it to DDPM \cite{ho2020denoising} on the CIFAR dataset, as shown in Table~\ref{tab:cifar10_results_trad}. 
Our EB-Diff-Train (EB) achieves 2.57$\times \sim$ 4.38$\times$ speedups over train-prune-finetune baselines with comparable or even lower FID ($\downarrow$0.12$\sim$$\uparrow$7.91). Additionally, our EB-Diff-Train (TA-EB) provides further improvements for extremely efficient training, achieving a speedup of up to 10.32$\times$ with comparable FID compared to train-prune-finetune baselines.
This set of experiments validates the effectiveness of our EB-Diff-Train method on CIFAR.

\begin{figure*}[t!]
\centering
\includegraphics[width=\linewidth, keepaspectratio]{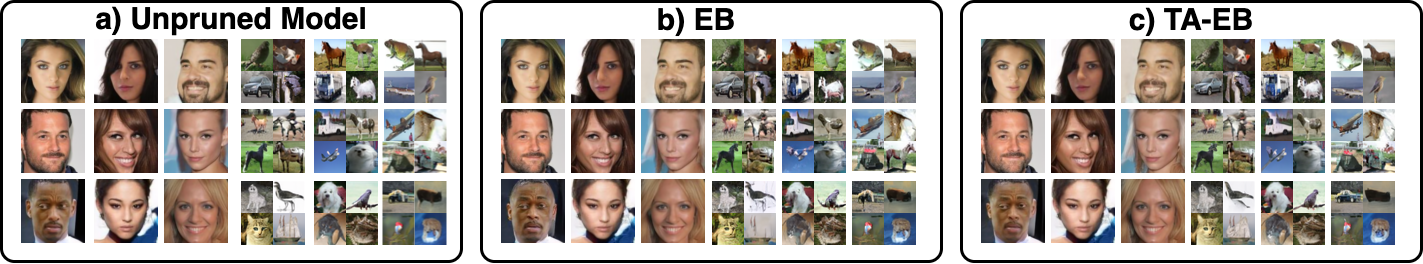}

\caption{Qualitative comparison of generated image results for CelebA (left 3$\times$ 3 section) and CIFAR-10 (right 6$\times$ 6 section) using the DDPM \cite{ho2020denoising} model. From left to right: Generations from the unpruned model, our Early-Bird (EB) model with a 50\% pruning rate, and our Timestep-Aware Early-Bird (TA-EB) 64\% average pruning rate.}
\vspace{-0.8em}
\label{fig:qualitative_comparison}
\end{figure*}

\begin{table}[t]
\centering
\caption{Results for CelebA using the DDPM \cite{ho2020denoising} model. The best FID/Speed-Up is \textbf{bolded}. ``Speed-Up'' is measured in the same fashion as Table \ref{tab:cifar10_results_trad}. The number of iterations for the \textbf{TA-EB} methods are recorded as the total number of iterations for the subnetwork of longest training time.}
\vspace{-0.5em}
\scriptsize 
\setlength{\tabcolsep}{3pt} 
\renewcommand{\arraystretch}{1.2} 
\resizebox{\linewidth}{!}{
\begin{tabular}{@{}lcccccc@{}}
\toprule
\multirow{2}[2]{*}{\tabincell{l}{\textbf{Pruning}\\\textbf{Metric}}} & \multicolumn{5}{c}{\textbf{DDPM @ CelebA (64$\times$64)}} \\ 
\cmidrule(lr){2-6}
& \textbf{\#Params} $\downarrow$ & \textbf{MACs} $\downarrow$ & \textbf{FID} $\downarrow$ & \textbf{Iters} & \textbf{Speed-Up} $\uparrow$\\
\midrule
Unpruned & 78.7M & 23.9G & 5.03 & 500K & 1.00$\times$\\
 \midrule 
 \multicolumn{6}{c}{\textbf{Baselines} w/ 50\% Pruning Rate} \\
 \cmidrule(lr){2-5}
\multirow{5}{*}{\shortstack[l]{Scratch \\ Random \\ Magnitude \\ Taylor \\ Diff-Pruning}} 
& \multirow{5}{*}{19.7M} & \multirow{5}{*}{3.4G} & 5.57 & 600K & 0.65$\times$\\
 & & & 6.52 & 100K & 0.91$\times$\\
 & & & 6.86 & 100K & 0.91$\times$\\
 & & & 6.09 & 100K & 0.91$\times$\\
 & & & 6.00 & 100K & 0.91$\times$\\
 \midrule 
 \multicolumn{6}{c}{\textbf{EB-Diff-Train (EB)} w/ 50\% Pruning Rate} \\
 \cmidrule(lr){2-5}
\multirow{3}{*}{\shortstack[l]{Magnitude \\ \textbf{Taylor} \\ Diff-Pruning}} 
& \multirow{3}{*}{19.7M} & \multirow{3}{*}{6.0G} & 5.53 & 600K & 1.89$\times$\\
 & & & \textbf{5.42} & 600K & \textbf{1.89$\times$}\\
 & & & 5.58 & 600K & 1.89$\times$\\
  \midrule 
 \multicolumn{6}{c}{\textbf{EB-Diff-Train (TA-EB)} w/ Avg. 64\% Pruning Rate} \\
 \cmidrule(lr){2-5}
 Magnitude & 16.5M & 5.2G & \textbf{5.41} & 200K & \textbf{4.38$\times$} \\
\bottomrule
\vspace{-3.0em}
\end{tabular}
}
\label{tab:celeba_results_trad}
\end{table}

\textbf{Results for CelebA.} Similarly, we extend our comparison to the CelebA dataset to evaluate face generation quality. As shown in Table~\ref{tab:celeba_results_trad}, our EB-Diff-Train consistently outperforms the baselines, with EB-Diff-Train (EB) achieving 2.08$\times$$\sim$2.91$\times$ speedups over train-prune-finetune baselines with comparable or even lower FID ($\downarrow$0.15$\sim$$\uparrow$1.44), and EB-Diff-Train (TA-EB) achieving 4.81$\times \sim$ 6.74$\times$ speedups with comparable or even lower FID ($\downarrow$0.16$\sim$$\uparrow$1.45).

\begin{table}[t]
    \centering
    \caption{Results for LSUN Church/Bedroom and ImageNet-1K. The best FID/Speed-Up is \textbf{bolded}. The number of iterations for the \textbf{TA-EB} methods are recorded as the total number of iterations for subnetwork of longest training time. DDPM~\cite{ho2020denoising} and LDM~\cite{rombach2021highresolution} are used for LSUN Church/Bedroom and ImageNet, respectively.}
    \vspace{-0.5em}
    \resizebox{\columnwidth}{!}{%
    \begin{tabular}{lccccc}
    \toprule
    \multicolumn{6}{c}{\textbf{DDPM @ LSUN Church (256$\times$256)}} \\
    \textbf{Method} & \textbf{\#Params $\downarrow$} & \textbf{MACs $\downarrow$} & \textbf{FID $\downarrow$} & \textbf{Iters} & \textbf{Speed-Up} $\uparrow$ \\
    \midrule
    Unpruned & 113.7M & 248.7G & 26.58 & 800K & 1.00$\times$\\
    \midrule
    Scratch & 28.5M & 62.4G & 54.18 & 800K & 0.68$\times$\\
    EB & 28.5M & 62.4G & \textbf{37.60} & 800K & 2.10$\times$\\
    TA-EB & 23.6M & 53.9G & 54.03 & 200K & \textbf{5.36$\times$} \\
    TA-EB & 23.6M & 53.9G & 46.12 & 300K & 3.57$\times$ \\
    \midrule
    \multicolumn{6}{c}{\textbf{DDPM @ LSUN Bedroom (256$\times$256) }} \\
    \textbf{Method} & \textbf{\#Params $\downarrow$} & \textbf{MACs$\downarrow$} & \textbf{FID $\downarrow$} & \textbf{Iters} & \textbf{Speed-Up} $\uparrow$\\
    \midrule
    Unpruned & 113.7M & 248.7G & 45.37 & 800K & 1.00$\times$ \\
    \midrule
    Scratch & 28.5M & 62.4G & 217.07 & 800K & 0.68$\times$\\
    EB & 28.5M & 62.4G & \textbf{81.60} & 800K  & 2.10$\times$\\
    TA-EB & 23.6M & 53.9G & 144.58 & 200K & \textbf{5.36$\times$} \\
    TA-EB & 23.6M & 53.9G & 135.20 & 300K & 3.57$\times$ \\
    \bottomrule
    \multicolumn{6}{c}{\textbf{LDM @ ImageNet-1K (256$\times$256)}} \\
    \textbf{Method} & \textbf{\#Params $\downarrow$} & \textbf{MACs $\downarrow$} & \textbf{FID $\downarrow$} & \textbf{Iters} & \textbf{Speed-Up} $\uparrow$\\
    \midrule
    Unpruned & 400.9M & 598.8G & 34.80 & 100K &1.00$\times$\\
    \midrule
    Scratch & 92.8M & 141.7G & 43.10 & 100K & 0.67$\times$ \\
    EB & 92.8M & 141.7G & \textbf{36.18} & 100K & 2.00$\times$\\
    TA-EB & 67.0M & 117.3G & 44.63 & 50K & \textbf{2.90$\times$} \\
    \bottomrule

    \end{tabular}
    }
    \label{tab:lsun_trad}
    \vspace{-1em}
\end{table}
\begin{table}[b!]
\centering
\caption{Results for CIFAR-10 32$\times$32 using the DDPM~\cite{ho2020denoising} model with 100 DDIM~\cite{song2020denoising} steps. The number of iterations for the \textbf{TA-EB} methods is recorded as the total number of iterations for the subnetwork with the longest training time. ``Speed-Up'' is calculated as the ratio of the training time for the unpruned model to that of the pruned model.}
\vspace{-0.5em}
\label{tab:speed_comparison}
\resizebox{\columnwidth}{!}{%
\begin{tabular}{lccc}
\toprule
\multicolumn{4}{c}{\textbf{DDPM @ CIFAR-10 (32$\times$32)}} \\
\midrule
\textbf{Method} & \textbf{Iterations} & \textbf{FID$\downarrow$} & \textbf{Speed-up} \\
\midrule
\multicolumn{4}{c}{\textit{Unpruned}} \\
\midrule
Unpruned & 300K & 5.63 & 1.00$\times$ \\
Unpruned + SpeeD & 300K & 4.76 & 1.00$\times$ \\
\midrule
\multicolumn{4}{c}{\textit{Comparisons w/ 50\% Pruning Rate}} \\
\midrule
Scratch & 300K & 11.29 & 0.71$\times$ \\
SpeeD & 300K & 9.01 & 2.45$\times$ \\
EB-Diff-Train (EB) & 300K & 9.32 & 2.45$\times$ \\
EB-Diff-Train (EB) + SpeeD & 300K & 8.89 & 2.45$\times$ \\
\midrule
\multicolumn{4}{c}{\textit{Comparisons w/ 50\% Avg. Pruning Rate}} \\
\midrule
EB-Diff-Train (TA-EB) & 100K & 9.34 & 4.33$\times$ \\
EB-Diff-Train (TA-EB) + SpeeD & 100K & 8.01 & 4.33$\times$ \\
\bottomrule
\end{tabular}%
}
\end{table}

\textbf{Results for LSUN and ImageNet.} To further validate the scalability of our proposed EB-Diff-Train, we evaluate it on larger datasets: LSUN Church, LSUN Bedroom, and ImageNet-1K. As shown in Table~\ref{tab:lsun_trad}, our EB-Diff consistently outperforms the ``scratch'' baseline 
on these higher-resolution datasets. Specifically, EB-Diff-Train (EB) achieves 2.99$\times \sim$ 3.09$\times$ speedups over train-prune-finetune baselines (i.e., ``scratch'') with comparable or even lower FID ($\downarrow$6.92$\sim$$\uparrow$135.47). EB-Diff-Train (TA-EB) achieves 4.33$\times \sim$ 7.88$\times$ speedups with comparable FID, demonstrating the method's effectiveness at scale.
Notably, to efficiently handle the larger dataset size, we used a ``pseudo-epoch'' of 1K steps to quickly identify EB tickets, and we chose magnitude pruning as the least computationally intensive method. Ablations on the pseudo-epoch choice are included in the supplementary materials.

\textbf{Comparison with SpeeD.}
Our EB-Diff-Train (EB) and EB-Diff-Train (TA-EB) strategies are efficient training methods designed for pruned models. Importantly, our approach is orthogonal to SOTA timestep resampling techniques and can be combined with those methods to achieve more benefits. 
To validate this, we compare and demonstrate how EB-Diff-Train (EB) and EB-Diff-Train (TA-EB) can be integrated with a SOTA timestep resampling method, SpeeD~\cite{wang2024closer}. In Table~\ref{tab:speed_comparison}, we compare these methods against both unpruned models and the ``scratch'' baseline introduced in Section~\ref{sec:exp_settings}. The results show that our EB-Diff-Train (EB) and EB-Diff-Train (TA-EB) methods can be applied on top of SpeeD effectively, leading to reduced FID scores of 2.40$\sim$3.28, as well as speedups of 3.45$\times$ and 6.10$\times$, respectively, compared to the ``scratch'' baseline.

\begin{table}[t!]
\centering
\caption{Comparison of different pruning rates for DDPM on the CIFAR-10 dataset in terms of FID scores and average MACs. ``Pruning Rate 1/2/3'' denotes the pruning rates applied to the model operating on timestep regions 1/2/3.}
\label{tab:avg_sparsity_timesteps_flops_fid}

\vspace{-0.5em}
\resizebox{\columnwidth}{!}{%
\begin{tabular}{c c c c c c c}
\toprule
\textbf{Avg. Pruning}  & \textbf{Pruning} & \textbf{Pruning} & \textbf{Pruning} & \multirow{2}{*}{\textbf{Avg MACs}$\downarrow$} & \multirow{2}{*}{\textbf{FID} $\downarrow$} \\ 
\textbf{Rate}  & \textbf{Rate 1} & \textbf{Rate 2} & \textbf{Rate 3} &  &  \\ 
\midrule
54.4\% & 30.0\% & 40.0\% & 70.0\% & 1.96G & 7.29 \\ 
60.0\%   & 30.0\% & 40.0\% & 80.0\% & 1.57G &7.48 \\ 
65.6\% & 30.0\% & 40.0\% & 90.0\% & 1.57G &17.61 \\ 
56.4\% & 30.0\% & 50.0\% & 70.0\% &  1.72G &7.53 \\ 
62.0\%   & 30.0\% & 50.0\% & 80.0\% & 1.33G &7.75 \\ 
58.4\% & 30.0\% & 60.0\% & 70.0\% & 1.72G & 7.50 \\ 
64.0\% & 30.0\% & 60.0\% & 80.0\% & 1.32G &7.69 \\ 
\bottomrule
\end{tabular}%
}
\end{table}

\vspace{-0.5em}
\subsection{Ablation Studies of Our EB-Diff-Train (TA-EB)}

\textbf{Ablation Study on Pruning Rates at Fixed Training Iterations.} 
To examine how different pruning rates, impact the quality-efficiency trade-offs, we summarize the corresponding comparison in Table~\ref{tab:avg_sparsity_timesteps_flops_fid}. In general, higher pruning rates tend to result in worse FID scores. However, even models with high average pruning rates (e.g., 64\% achieved with 30\%/60\%/80\% pruning rates across regions) maintain competitive FID scores compared to our prior EB tickets with 50\% pruning rate. This suggests that highly pruned models (e.g., 80\%) can effectively handle most timesteps, requiring lower pruning rates (e.g., 30\%) for only a small subset of timesteps.

\textbf{Ablation Study on Training Iterations per Region.} To examine how training iterations in each region affect the resulting FID scores, we vary the iterations for each region under the 30\%/60\%/80\% pruning rates scheme and summarize the results in Table~\ref{tab:hybrid_sparsity_hybrid_steps}. Our observations reveal a non-monotonic relationship between training iterations and FID, indicating that optimal FID scores can be achieved with a relatively lower number of training iterations, as highlighted in bold in Table~\ref{tab:hybrid_sparsity_hybrid_steps}.

\begin{table}[t!]
\centering
\caption{Comparison of using different training iterations in each region in terms of FID scores. The best FID is \textbf{bolded}.}
\vspace{-1em}

\resizebox{\columnwidth}{!}{%
\begin{tabular}{cccc}
\toprule
\textbf{80\% Model \# Iters} & \textbf{60\% Model \# Iters} & \textbf{30\% Model \# Iters} & \textbf{FID} \(\downarrow\) \\ 
\midrule
200K & 200K & 200K & 7.69 \\
200K & 200K & 300K & 7.56 \\
200K & 300K & 300K & 7.64\\
\textbf{300K} & \textbf{300K} & \textbf{300K} & \textbf{7.40}\\
300K & 300K & 400K & 7.49\\
300K & 400K & 400K & 7.60\\
400K & 400K & 400K & 7.49\\
\bottomrule
\end{tabular}%
}
\label{tab:hybrid_sparsity_hybrid_steps}
\vspace{-1em}
\end{table}

\vspace{-0.5em}
\subsection{Qualitative Visual Comparison}
\vspace{-0.5em}
To assess qualitative performance, we visualize the generated images from an unpruned model, our EB ticket, and our TA-EB ticket in Figure \ref{fig:qualitative_comparison}. This comparison shows that the observed 2.08$\times$ to 6.74$\times$ speedups over training the unpruned dense model are achieved with minimal perceived differences between the unpruned and pruned models.
\vspace{-0.8em}
\section{Conclusion}
\vspace{-0.5em}

In this work, we propose EB-Diff-Train, an efficient DM training framework leveraging both traditional and timestep-aware EB tickets. Our approach identifies traditional EB tickets in DMs and extends this concept by exploring the varying importance of timestep regions, introducing timestep-aware EB tickets with distinct pruning rates for different regions. These specialized subnetworks are trained in parallel and combined during inference. Experiments demonstrate EB-Diff-Train achieves 2.9$\times$ to 5.8$\times$ faster training than dense models and up to 10.3$\times$ speedup over train-prune-finetune baselines.

\vspace{-0.5em}
\section*{Acknowledgment}
\vspace{-0.2em}

This work is supported by the Department of Health and Human Services Advanced Research Projects Agency for Health (ARPA-H) under Award Number AY1AX000003, and CoCoSys, one of the seven centers in JUMP 2.0, a Semiconductor Research Corporation (SRC) program sponsored by DARPA. This material is based upon work by the National Science Foundation Graduate Research Fellowship under Grant No. DGE-2039655. Any opinions, findings, and conclusions or recommendations expressed in this material are those of the author(s) and do not necessarily reflect the views of the National Science Foundation.

{
    \small
    \bibliographystyle{ieeenat_fullname}
    \bibliography{main}

\begin{thebibliography}{41}
\providecommand{\natexlab}[1]{#1}
\providecommand{\url}[1]{\texttt{#1}}
\expandafter\ifx\csname urlstyle\endcsname\relax
  \providecommand{\doi}[1]{doi: #1}\else
  \providecommand{\doi}{doi: \begingroup \urlstyle{rm}\Url}\fi

\bibitem[Balaji et~al.(2022)Balaji, Nah, Huang, Vahdat, Song, Zhang, Kreis, Aittala, Aila, Laine, Catanzaro, Karras, and Liu]{balaji2023ediffitexttoimagediffusionmodels}
Yogesh Balaji, Seungjun Nah, Xun Huang, Arash Vahdat, Jiaming Song, Qinsheng Zhang, Karsten Kreis, Miika Aittala, Timo Aila, Samuli Laine, Bryan Catanzaro, Tero Karras, and Ming-Yu Liu.
\newblock {eDiff-I}: Text-to-image diffusion models with an ensemble of expert denoisers.
\newblock \emph{arXiv preprint arXiv:2211.01324}, 2022.

\bibitem[Chen et~al.(2021)Chen, Cheng, Wang, Gan, Wang, and Liu]{chen2020earlybert}
Xiaohan Chen, Yu Cheng, Shuohang Wang, Zhe Gan, Zhangyang Wang, and Jingjing Liu.
\newblock {EarlyBERT}: Efficient {BERT} training via early-bird lottery tickets.
\newblock In \emph{Annual Meeting of the Association for Computational Linguistics and International Joint Conference on Natural Language Processing}, pages 2195--2207. ACL, 2021.

\bibitem[Choi et~al.(2022)Choi, Lee, Shin, Kim, Kim, and Yoon]{Choi_2022_CVPR}
Jooyoung Choi, Jungbeom Lee, Chaehun Shin, Sungwon Kim, Hyunwoo Kim, and Sungroh Yoon.
\newblock Perception prioritized training of diffusion models.
\newblock In \emph{Computer Vision and Pattern Recognition}, pages 11472--11481, 2022.

\bibitem[Deng et~al.(2009)Deng, Dong, Socher, Li, Li, and Fei{-}Fei]{deng2009imagenet}
Jia Deng, Wei Dong, Richard Socher, Li{-}Jia Li, Kai Li, and Li Fei{-}Fei.
\newblock {ImageNet: A large-scale hierarchical image database}.
\newblock In \emph{Computer Vision and Pattern Recognition}, pages 248--255, 2009.

\bibitem[Dhariwal and Nichol(2021)]{diffModelsBeatGANs}
Prafulla Dhariwal and Alexander~Quinn Nichol.
\newblock Diffusion models beat gans on image synthesis.
\newblock In \emph{Advances in Neural Information Processing Systems}, pages 8780--8794, 2021.

\bibitem[Ding et~al.(2024)Ding, Zhang, Wu, and Tu]{ding2023patcheddenoisingdiffusionmodels}
Zheng Ding, Mengqi Zhang, Jiajun Wu, and Zhuowen Tu.
\newblock Patched denoising diffusion models for high-resolution image synthesis.
\newblock In \emph{International Conference on Learning Representations}, 2024.

\bibitem[Fang et~al.(2023)Fang, Ma, and Wang]{diffPruning}
Gongfan Fang, Xinyin Ma, and Xinchao Wang.
\newblock Structural pruning for diffusion models.
\newblock In \emph{Advances in Neural Information Processing Systems}, pages 16716--16728, 2023.

\bibitem[Frankle and Carbin(2019)]{frankle2018the}
Jonathan Frankle and Michael Carbin.
\newblock The lottery ticket hypothesis: Finding sparse, trainable neural networks.
\newblock In \emph{International Conference on Learning Representations}, 2019.

\bibitem[Go et~al.(2023{\natexlab{a}})Go, , Lee, Lee, Oh, Moon, and Choi]{NEURIPS2023_56a7b9a0}
Hyojun Go, , Yunsung Lee, Seunghyun Lee, Shinhyeok Oh, Hyeongdon Moon, and Seungtaek Choi.
\newblock Addressing negative transfer in diffusion models.
\newblock In \emph{Advances in Neural Information Processing Systems}, pages 27199--27222, 2023{\natexlab{a}}.

\bibitem[Go et~al.(2023{\natexlab{b}})Go, Lee, Kim, Lee, Jeong, Lee, and Choi]{go2023towards}
Hyojun Go, Yunsung Lee, Jin-Young Kim, Seunghyun Lee, Myeongho Jeong, Hyun~Seung Lee, and Seungtaek Choi.
\newblock Towards practical plug-and-play diffusion models.
\newblock In \emph{Computer Vision and Pattern Recognition}, pages 1962--1971, 2023{\natexlab{b}}.

\bibitem[Goodfellow et~al.(2014)Goodfellow, Pouget-Abadie, Mirza, Xu, Warde-Farley, Ozair, Courville, and Bengio]{goodfellow2014generativeadversarialnetworks}
Ian~J. Goodfellow, Jean Pouget-Abadie, Mehdi Mirza, Bing Xu, David Warde-Farley, Sherjil Ozair, Aaron Courville, and Yoshua Bengio.
\newblock Generative adversarial nets.
\newblock In \emph{International Conference on Neural Information Processing Systems}, page 2672–2680, 2014.

\bibitem[Han et~al.(2015)Han, Pool, Tran, and Dally]{hanMagnitudePrune}
Song Han, Jeff Pool, John Tran, and William Dally.
\newblock Learning both weights and connections for efficient neural network.
\newblock In \emph{Advances in Neural Information Processing Systems}, pages 1135--1143, 2015.

\bibitem[Hang et~al.(2023)Hang, Gu, Li, Bao, Chen, Hu, Geng, and Guo]{hang2024efficientdiffusiontrainingminsnr}
Tiankai Hang, Shuyang Gu, Chen Li, Jianmin Bao, Dong Chen, Han Hu, Xin Geng, and Baining Guo.
\newblock Efficient diffusion training via min-snr weighting strategy.
\newblock In \emph{International Conference on Computer Vision}, pages 7407--7417, 2023.

\bibitem[Heusel et~al.(2017)Heusel, Ramsauer, Unterthiner, Nessler, and Hochreiter]{fid_score}
Martin Heusel, Hubert Ramsauer, Thomas Unterthiner, Bernhard Nessler, and Sepp Hochreiter.
\newblock Gans trained by a two time-scale update rule converge to a local nash equilibrium.
\newblock In \emph{Advances in Neural Information Processing Systems}, pages 6626--6637, 2017.

\bibitem[Ho et~al.(2020)Ho, Jain, and Abbeel]{ho2020denoising}
Jonathan Ho, Ajay Jain, and Pieter Abbeel.
\newblock Denoising diffusion probabilistic models.
\newblock In \emph{Advances in Neural Information Processing Systems}, pages 6840--6851, 2020.

\bibitem[Hyv{{\"a}}rinen(2005)]{hyvarianScoreMatching}
Aapo Hyv{{\"a}}rinen.
\newblock Estimation of non-normalized statistical models by score matching.
\newblock \emph{Journal of Machine Learning Research}, 6\penalty0 (24):\penalty0 695--709, 2005.

\bibitem[Kalibhat et~al.(2021)Kalibhat, Balaji, and Feizi]{gan_eb}
Neha~Mukund Kalibhat, Yogesh Balaji, and Soheil Feizi.
\newblock Winning lottery tickets in deep generative models.
\newblock In \emph{AAAI Conference on Artificial Intelligence}, pages 8038--8046, 2021.

\bibitem[Kim et~al.(2024)Kim, Song, Castells, and Choi]{kim2023bksdmlightweightfastcheap}
Bo{-}Kyeong Kim, Hyoung{-}Kyu Song, Thibault Castells, and Shinkook Choi.
\newblock {BK-SDM:} {A} lightweight, fast, and cheap version of stable diffusion.
\newblock In \emph{European Conference on Computer Vision}, pages 381--399, 2024.

\bibitem[Kingma and Welling(2014)]{kingma2022autoencodingvariationalbayes}
Diederik~P. Kingma and Max Welling.
\newblock Auto-encoding variational bayes.
\newblock In \emph{International Conference on Learning Representations}, 2014.

\bibitem[Krizhevsky et~al.(2009)Krizhevsky, Hinton, et~al.]{cifar10}
Alex Krizhevsky, Geoffrey Hinton, et~al.
\newblock Learning multiple layers of features from tiny images.
\newblock Technical Report TR-2009, University of Toronto, Toronto, ON, Canada, 2009.

\bibitem[Lee et~al.(2024)Lee, Kim, Go, Jeong, Oh, and Choi]{lee2023multiarchitecturemultiexpertdiffusionmodels}
Yunsung Lee, JinYoung Kim, Hyojun Go, Myeongho Jeong, Shinhyeok Oh, and Seungtaek Choi.
\newblock Multi-architecture multi-expert diffusion models.
\newblock In \emph{AAAI Conference on Artificial Intelligence}, pages 13427--13436, 2024.

\bibitem[Liu et~al.(2023)]{oms_dpm}
Liu et~al.
\newblock {OMS}-{DPM}: Optimizing the model schedule for diffusion probabilistic models.
\newblock In \emph{International Conference on Machine Learning}, pages 21915--21936, 2023.

\bibitem[Liu et~al.(2015)Liu, Luo, Wang, and Tang]{celeba}
Ziwei Liu, Ping Luo, Xiaogang Wang, and Xiaoou Tang.
\newblock Deep learning face attributes in the wild.
\newblock In \emph{International Conference on Computer Vision}, pages 3730--3738, 2015.

\bibitem[Lu et~al.(2022)Lu, Zhou, Bao, Chen, Li, and Zhu]{lu2022dpmsolverfastodesolver}
Cheng Lu, Yuhao Zhou, Fan Bao, Jianfei Chen, Chongxuan Li, and Jun Zhu.
\newblock Dpm-solver: {A} fast {ODE} solver for diffusion probabilistic model sampling in around 10 steps.
\newblock In \emph{Advances in Neural Information Processing Systems}, pages 5775--5787, 2022.

\bibitem[Molchanov et~al.(2019)Molchanov, Mallya, Tyree, Frosio, and Kautz]{Molchanov_2019_CVPR}
Pavlo Molchanov, Arun Mallya, Stephen Tyree, Iuri Frosio, and Jan Kautz.
\newblock Importance estimation for neural network pruning.
\newblock In \emph{Computer Vision and Pattern Recognition}, 2019.

\bibitem[OpenAI(2022)]{openai2022dalle2}
OpenAI.
\newblock {DALL{\textperiodcentered}E 2}.
\newblock \url{https://openai.com/dall-e-2}, 2022.
\newblock [Online; accessed 12-November-2024].

\bibitem[Phung et~al.(2023)Phung, Dao, and Tran]{phung2023waveletdiffusionmodelsfast}
Hao Phung, Quan Dao, and Anh Tran.
\newblock Wavelet diffusion models are fast and scalable image generators.
\newblock In \emph{Computer Vision and Pattern Recognition}, pages 10199--10208, 2023.

\bibitem[Rombach et~al.(2022)Rombach, Blattmann, Lorenz, Esser, and Ommer]{rombach2021highresolution}
Robin Rombach, Andreas Blattmann, Dominik Lorenz, Patrick Esser, and Bj{\"{o}}rn Ommer.
\newblock High-resolution image synthesis with latent diffusion models.
\newblock In \emph{Computer Vision and Pattern Recognition}, pages 10674--10685, 2022.

\bibitem[Sohl-Dickstein et~al.(2015)Sohl-Dickstein, Weiss, Maheswaranathan, and Ganguli]{deepUnsupervisedLearningNonequilibrium}
Jascha Sohl-Dickstein, Eric Weiss, Niru Maheswaranathan, and Surya Ganguli.
\newblock Deep unsupervised learning using nonequilibrium thermodynamics.
\newblock In \emph{International Conference on Machine Learning}, pages 2256--2265, 2015.

\bibitem[Song et~al.(2021{\natexlab{a}})Song, Meng, and Ermon]{song2020denoising}
Jiaming Song, Chenlin Meng, and Stefano Ermon.
\newblock Denoising diffusion implicit models.
\newblock In \emph{International Conference on Learning Representations}, 2021{\natexlab{a}}.

\bibitem[Song and Ermon(2019)]{song2019generative}
Yang Song and Stefano Ermon.
\newblock Generative modeling by estimating gradients of the data distribution.
\newblock In \emph{Advances in Neural Information Processing Systems}, pages 11895--11907, 2019.

\bibitem[Song et~al.(2019)Song, Garg, Shi, and Ermon]{song2019sliced}
Yang Song, Sahaj Garg, Jiaxin Shi, and Stefano Ermon.
\newblock Sliced score matching: {A} scalable approach to density and score estimation.
\newblock In \emph{Conference on Uncertainty in Artificial Intelligence}, pages 574--584, 2019.

\bibitem[Song et~al.(2021{\natexlab{b}})Song, Durkan, Murray, and Ermon]{song2021scorebased}
Yang Song, Conor Durkan, Iain Murray, and Stefano Ermon.
\newblock Maximum likelihood training of score-based diffusion models.
\newblock In \emph{Advances in Neural Information Processing Systems}, pages 1415--1428, 2021{\natexlab{b}}.

\bibitem[Wang et~al.(2024)Wang, Shi, Zhou, Li, Yuan, Shang, Peng, Zhang, and You]{wang2024closer}
Kai Wang, Mingjia Shi, Yukun Zhou, Zekai Li, Zhihang Yuan, Yuzhang Shang, Xiaojiang Peng, Hanwang Zhang, and Yang You.
\newblock A closer look at time steps is worthy of triple speed-up for diffusion model training.
\newblock \emph{arXiv preprint arXiv:2405.17403}, 2024.

\bibitem[Wang et~al.(2023)Wang, Jiang, Zheng, Wang, He, Wang, Chen, and Zhou]{wang2023patchdiffusionfasterdataefficient}
Zhendong Wang, Yifan Jiang, Huangjie Zheng, Peihao Wang, Pengcheng He, Zhangyang Wang, Weizhu Chen, and Mingyuan Zhou.
\newblock Patch diffusion: Faster and more data-efficient training of diffusion models.
\newblock In \emph{Advances in Neural Information Processing Systems}, pages 72137--72154, 2023.

\bibitem[Xue et~al.(2023)Xue, Song, Guo, Liu, Zong, Liu, and Luo]{xue2024raphaeltexttoimagegenerationlarge}
Zeyue Xue, Guanglu Song, Qiushan Guo, Boxiao Liu, Zhuofan Zong, Yu Liu, and Ping Luo.
\newblock {RAPHAEL:} text-to-image generation via large mixture of diffusion paths.
\newblock In \emph{Advances in Neural Information Processing Systems}, pages 41693--41706, 2023.

\bibitem[You et~al.(2020)You, Li, Xu, Fu, Wang, Chen, Baraniuk, Wang, and Lin]{earlyBird}
Haoran You, Chaojian Li, Pengfei Xu, Yonggan Fu, Yue Wang, Xiaohan Chen, Richard~G. Baraniuk, Zhangyang Wang, and Yingyan Lin.
\newblock Drawing early-bird tickets: Toward more efficient training of deep networks.
\newblock In \emph{International Conference on Learning Representations}, 2020.

\bibitem[You et~al.(2022)You, Lu, Zhou, Fu, and Lin]{you2021earlybirdgcnsgraphnetworkcooptimization}
Haoran You, Zhihan Lu, Zijian Zhou, Yonggan Fu, and Yingyan Lin.
\newblock Early-bird gcns: Graph-network co-optimization towards more efficient {GCN} training and inference via drawing early-bird lottery tickets.
\newblock In \emph{AAAI Conference on Artificial Intelligence}, pages 8910--8918, 2022.

\bibitem[Yu et~al.(2015)Yu, Seff, Zhang, Song, Funkhouser, and Xiao]{lsun}
Fisher Yu, Ari Seff, Yinda Zhang, Shuran Song, Thomas Funkhouser, and Jianxiong Xiao.
\newblock Lsun: Construction of a large-scale image dataset using deep learning with humans in the loop.
\newblock \emph{arXiv preprint arXiv:1506.03365}, 2015.

\bibitem[Zhang et~al.(2024{\natexlab{a}})Zhang, Li, Chen, Xie, and Lu]{zhang2024laptopdifflayerpruningnormalized}
Dingkun Zhang, Sijia Li, Chen Chen, Qingsong Xie, and Haonan Lu.
\newblock Laptop-diff: Layer pruning and normalized distillation for compressing diffusion models.
\newblock \emph{arXiv preprint arXiv:2404.11098}, 2024{\natexlab{a}}.

\bibitem[Zhang et~al.(2024{\natexlab{b}})Zhang, Lu, Alkhouri, Ravishankar, Song, and Qu]{Zhang_2024_CVPR}
Huijie Zhang, Yifu Lu, Ismail Alkhouri, Saiprasad Ravishankar, Dogyoon Song, and Qing Qu.
\newblock Improving training efficiency of diffusion models via multi-stage framework and tailored multi-decoder architecture.
\newblock In \emph{Computer Vision and Pattern Recognition}, pages 7372--7381, 2024{\natexlab{b}}.

\end{thebibliography}
}


\end{document}


\clearpage
\appendix
\setcounter{page}{1}
\maketitlesupplementary

We provide additional supporting experiments and visualizations in this supplementary material.
Specifically, Section~\ref{sec:existence_eb_tickets} presents further empirical evidence for the existence of Early-Bird (EB) tickets in diffusion models across four additional dataset-model pairs. 
Section~\ref{sec:existence_ta_eb_tickets} offers additional empirical evidence of Timestep-Aware Early-Bird (TA-EB) tickets using four additional dataset-model pairs. 
Section~\ref{sec:ablation_pseudo_epoch} supplies an ablation study to analyze the impact of varying the number of iterations within a pseudo-epoch.
Section~\ref{sec:additional_generations} showcases additional image generations from the CelebA~\cite{celeba} dataset, further illustrating the qualitative performance of our proposed methods. Section~\ref{sec:dit_exps} presents additional experimental results using the Diffusion Transformer (DiT~\cite{Peebles2022DiT}) architecture, demonstrating the versatility of our approach when applied to this widely-used architecture. Section \ref{sec:regions_ablation} provides experimental results for different number of timestep regions, showcasing that our method is robust to number or regions. Section~\ref{sec:iterations_ablation} presents an ablation study on the number of iterations chosen for the TA-EB tickets, explaining how we determined the appropriate iteration count for our models.

\section{More Results for Finding EB Tickets}
\label{sec:existence_eb_tickets}

\begin{figure*}[t!]
    \centering
    \includegraphics[width=0.8\textwidth]{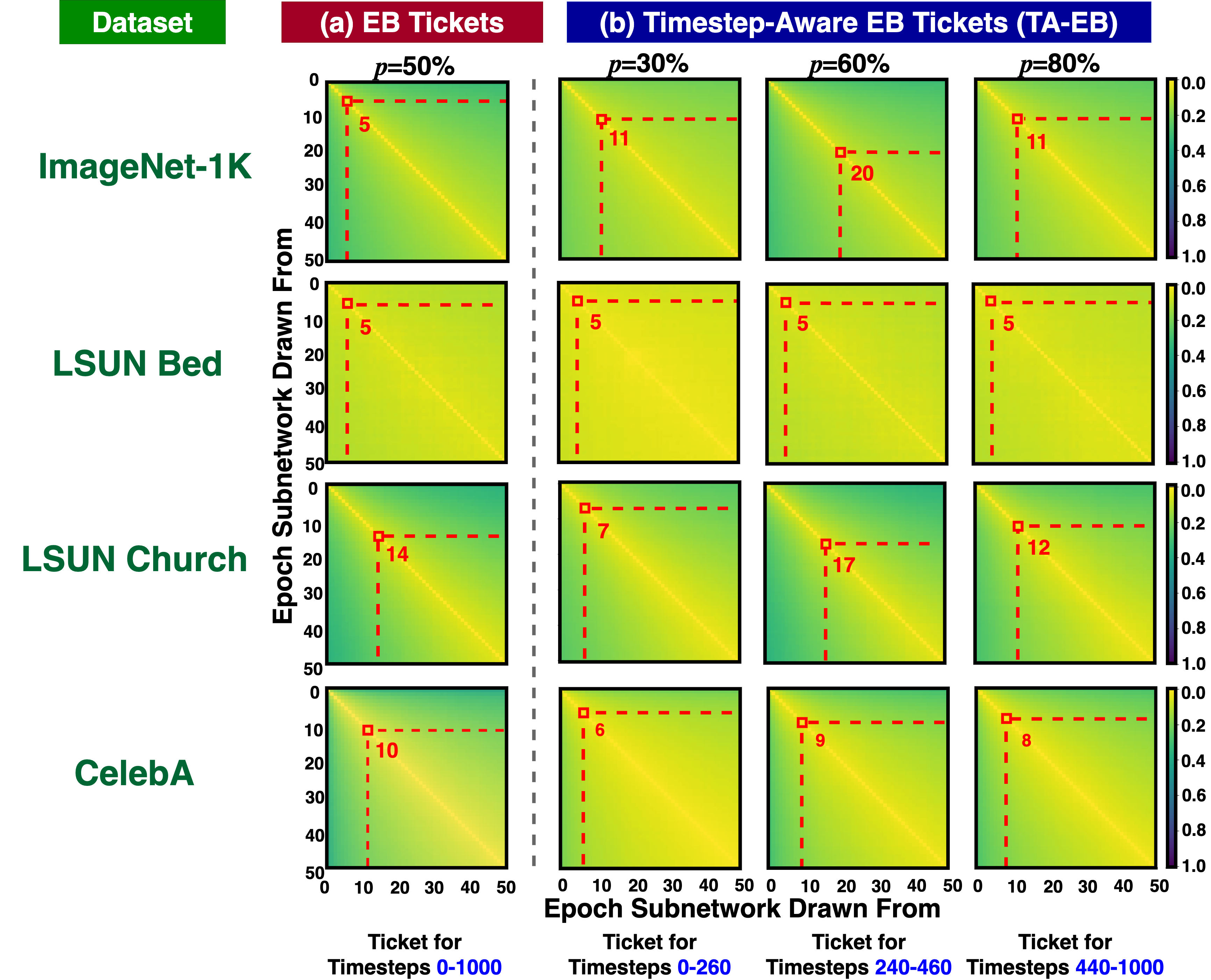} 
    \caption{(a) Visualizations of pairwise Hamming distance matrices of EB tickets for the CelebA~\cite{celeba}, LSUN Church~\cite{lsun}, LSUN Bedroom~\cite{lsun}, and ImageNet-1K~\cite{deng2009imagenet} datasets when using structural magnitude pruning at pruning rate of 50\%. (b) Visualizations of Hamming distance matrices of TA-EB tickets when using structural magnitude pruning with pruning rates of 30\%, 60\%, and 80\% across timestep periods of 0-260, 240-460, and 440-1000, respectively.} 
    \label{fig:eb_taeb_heatmaps} 
\end{figure*}

To provide further empirical evidence for identifying EB tickets in diffusion models, in Figure~\ref{fig:eb_taeb_heatmaps} (a) we present visualizations of pairwise mask distances across four dataset-model pairs: CelebA~\cite{celeba}, LSUN Church~\cite{lsun}, and LSUN Bedroom~\cite{lsun}, utilizing Denoising Diffusion Probabilistic Models (DDPMs)~\cite{ho2020denoising}, as well as ImageNet-1K~\cite{deng2009imagenet}, using the Latent Diffusion Model (LDM)~\cite{rombach2021highresolution}. For all settings, we apply magnitude-based structural pruning with a pruning rate of 50\%.

Due to the large number of training samples, we adopt a ``pseudo-epoch'' of 1K steps/iterations to expedite the identification of EB tickets in the case of LSUN Church, LSUN Bedroom, and ImageNet-1K.
The $(i,j)$-th element represents the Hamming distances between the pruned subnetworks extracted at the $i$-th and $j$-th pseudo-epochs.

Lighter colors correspond to lower inter-mask Hamming distances, darker colors indicate higher distances. The epoch/pseudo-epoch where the EB ticket is identified is marked in red font. Unless otherwise stated, we use a convergence threshold of $\eta=0.1$ and a FIFO queue of length 5 for these visualizations. The results demonstrate that EB tickets are consistently identified during the early stages of training.
This set of visualizations further empirically confirms the existence of EB tickets in diffusion model training.

\begin{figure}[t!]
    \centering
    \includegraphics[width=\columnwidth]{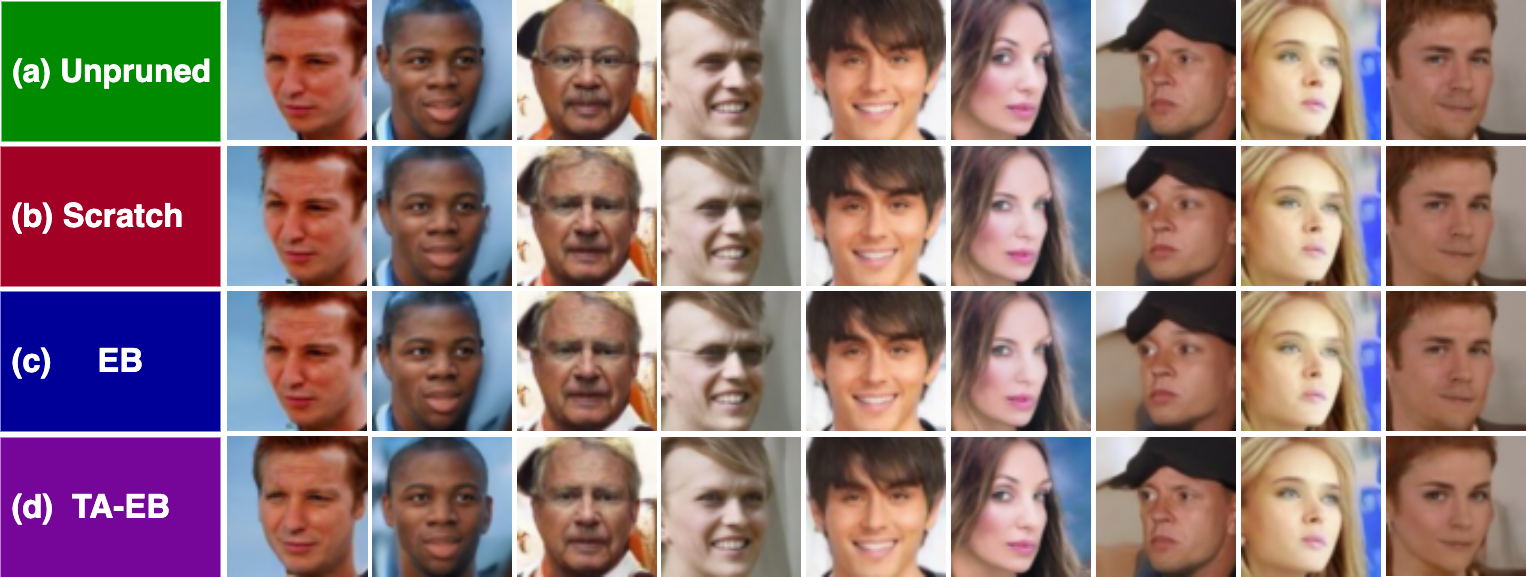}
    \vspace{-1.5em}
    \caption{More generations from the CelebA~\cite{celeba} dataset. a) Generations from the unpruned model. b) Generations from the ``Scratch" model with 50\% pruning rate, following the procedure in Section 5.1 of the manuscript. c) Generations from our EB-Diff-Train (EB) method with 50\% pruning rate. d) Generations from our EB-Diff-Train (TA-EB) method with 64\% average pruning rate. }
    \label{fig:extra_generations_celeba}
    \vspace{-1.5em}
\end{figure}

\section{More Results for Finding TA-EB Tickets}
\label{sec:existence_ta_eb_tickets}

To provide further empirical evidence for identifying TA-EB tickets, we present visualizations of pairwise mask distances across four dataset-model pairs in Figure~\ref{fig:eb_taeb_heatmaps} (b): CelebA~\cite{celeba}, LSUN Church~\cite{lsun}, and LSUN Bedroom~\cite{lsun}, all utilizing the DDPM~\cite{ho2020denoising}, as well as ImageNet-1K~\cite{deng2009imagenet}, using the LDM~\cite{rombach2021highresolution}.
For these experiments, we use magnitude-based structural pruning with pruning rates of 30\%, 60\%, and 80\%, respectively. 
Similar to the case of EB tickets (Section~\ref{sec:existence_eb_tickets}), we adopt a ``pseudo-epoch'' approach for LSUN Church, LSUN Bedroom, and ImageNet-1K datasets to identify TA-EB tickets more efficiently. Unless otherwise stated, we use a pseudo-epoch of 1000 steps/iterations.

The $(i,j)$-th element of each matrix represents the Hamming distance between subnetworks pruned at the $i$-th and $j$-th pseudo-epochs across the designated timestep regions. Lighter colors correspond to lower inter-mask Hamming distances, darker colors indicate higher distances. The epoch/pseudo-epoch where the TA-EB ticket is identified is marked in red font. Unless otherwise stated, we use a convergence threshold of $\eta=0.1$ and a FIFO queue of length 5 for these visualizations. 
The results show that TA-EB tickets are consistently identified during the early stages of training, further confirming their existence in diffusion model training.

\section{Ablation Study on Pseudo-Epoch Choices}
\label{sec:ablation_pseudo_epoch}
\begin{table}[b]
\centering
\caption{Ablation study on the choice of iterations in one pseudo-epoch (PE) for the LSUN Church 256$\times$256 dataset using the DDPM~\cite{ho2020denoising} model with 100 DDIM timesteps. 
}
\vspace{-0.5em}
\scriptsize 
\setlength{\tabcolsep}{1.5pt} 
\resizebox{\columnwidth}{!}{ 
\begin{tabular}{@{}lc|cc|cc@{}}
\toprule
\multicolumn{6}{c}{\textbf{LSUN Church 256$\times$256}} \\
\midrule
\textbf{Method} & \textbf{Iters per PE} & \textbf{\#Params} $\downarrow$ & \textbf{MACs} $\downarrow$ & \textbf{FID} $\downarrow$ & \textbf{Speed-Up} $\uparrow$\\
\midrule
Unpruned & - & 113.7M & 248.7G & 26.58 & 1.00$\times$\\
\midrule
Scratch & - & \multirow{7}{*}{28.5M} & \multirow{7}{*}{62.4G} & 54.18 & 0.68$\times$\\
EB-Diff-Train (EB) & 100 & & & 37.89 & 2.10$\times$\\
EB-Diff-Train (EB) & 300 & & & 40.30 & 2.10$\times$\\
EB-Diff-Train (EB) & 500 & & & 38.05 & 2.10$\times$\\
EB-Diff-Train (EB) & 1000 & & & 37.60 & 2.10$\times$\\
EB-Diff-Train (EB) & 3000 & & & 34.40 & 2.10$\times$\\
EB-Diff-Train (EB) & 5000 & & & 36.90 & 2.10$\times$\\
\bottomrule
\end{tabular}
}
\label{tab:sm_ablation_pseudo_epoch}
\end{table}

We conduct ablation studies to evaluate the impact of the number of iterations per pseudo-epoch on drawn tickets' performance. We employ magnitude-based structural pruning to prune DDPMs~\cite{ho2020denoising}, using the LSUN Church dataset~\cite{lsun} as a representative case. All training hyperparameters are consistent with those in~\cite{diffPruning}, and we use 100 DDIM~\cite{song2020denoising} timesteps to generate images. 
Table~\ref{tab:sm_ablation_pseudo_epoch} presents results examining the effect of varying the number of iterations within a single pseudo-epoch, ranging from 100 to 5K. To provide a robust comparison, we also include results for a pruned network trained from scratch, which involves pruning a fully trained network to establish its connectivity and subsequently reinitializing it with random weights for training. As prior work~\cite{diffPruning} indicates, training pruned networks from scratch is highly competitive, making this a critical baseline for analysis.
We observe that the EB tickets identified in our experiments are not sensitive to the choice of pseudo-epoch, as the pseudo-epoch choice marginally impacts the final image generation quality. In all cases, we achieve a 13.88$\sim$19.78 lower FID as well as 3.09$\times$ speedups compared to the ``scratch'' baseline.

\section{Additional Generations}
\label{sec:additional_generations}
To further qualitatively show the efficacy of our EB-Diff-Train (EB) and EB-Diff-Train (TA-EB) methods, we show additional generations from the CelebA~\cite{celeba} dataset in Fig.~\ref{fig:extra_generations_celeba}. We compare our EB and TA-EB methods against two baselines: (1) the original unpruned network (``Unpruned") and (2) a 50\% magnitude-based structurally pruned network trained from scratch following the methodology outlined in Section \ref{sec:existence_ta_eb_tickets} (``Scratch''). The generations show that our EB and TA-EB tickets can generate images of high quality while being up to 6.74$\times$ faster than the ``Scratch" baseline.

\section{Experiments Using the DiT}
\label{sec:dit_exps}
To highlight that our method can be applied to a variety of model architectures, we include results from the popular Diffusion Transformer (DiT \cite{Peebles2022DiT}) architecture in Table~\ref{tab:ablation_num_models} using the ImageNet-1K~\cite{deng2009imagenet} dataset. Following the methodology of Section \ref{sec:existence_ta_eb_tickets}, we compare against an unpruned network (``Unpruned'') and a magnitude-based structurally pruned network trained from scratch (``Scratch''). The results show that our EB-Diff-Train (EB) and EB-Diff-Train (TA-EB) methods can be successfully applied to the DiT architecture, yielding reductions in FID scores of $0.13\sim19.07$ while also improving training speed by $3.65\times\sim4.32\times$ respectively, as compared to the ``Scratch" baseline.

\section{Ablation Study on Number of Regions}
\label{sec:regions_ablation}
To showcase that our method is robust to the number of regions selected, we test 2, 3, and 4 regions, following the same settings as the submitted manuscript, in Table~\ref{tab:ablation_num_models}. For the 2 region case, we merged the first two timestep regions; for the 4 region case, we split the last region in half. The pruning rate was adjusted to maintain a timestep-weighted average of $\sim64\%$. From this ablation study, we see that our method is robust to the number of regions selected.

\begin{table}[t!]
    \centering
    \resizebox{\columnwidth}{!}{%
    \begin{tabular}{lccccc}
    \toprule
    \multicolumn{6}{c}{\textbf{DiT @ ImageNet-1K 256$\times$256}} \\
    \textbf{Method} & \textbf{\#Params $\downarrow$} & \textbf{MACs $\downarrow$} & \textbf{FID $\downarrow$} & \textbf{Iters} & \textbf{Speed-Up} $\uparrow$\\
    \midrule
    Unpruned & 675.1M & 118.7G & 28.12 & 100K &1.00$\times$\\
    \midrule
    Scratch & 337.5M & 57.0G & 58.23 & 100K & 0.72$\times$ \\
    \textbf{EB} & 337.5M & 57.0G & \textbf{33.16} & 100K & 2.63$\times$\\
    \textbf{TA-EB} & 242.1M & 40.7G & 52.10 & 40K & \textbf{3.11}$\times$\\
    \bottomrule
    
    \end{tabular}
    }
    \label{tab:dit_imagenet}
    \vspace{-1.2em}
\end{table}

\begin{table}[b!]
    \caption{Results for CIFAR-10 32$\times$32 using the DDPM~\cite{ho2020denoising} model with 100 DDIM steps under different region numbers. The number of iterations for
the TA-EB methods are recorded as the total number of iterations
for the subnetwork of longest training time}
    \label{tab:ablation_num_models}
    \centering
    \vspace{-1em}
    \resizebox{\columnwidth}{!}{%
    \begin{tabular}{lcccccc}
    \toprule
    \multicolumn{7}{c}{\textbf{DDPM @ CIFAR10 32$\times$32}} \\
    \textbf{\#Regions} & \textbf{Avg. Pruning Rate} & \textbf{\#Params $\downarrow$} & \textbf{MACs $\downarrow$} & \textbf{FID $\downarrow$} & \textbf{Iters} & \textbf{Speed-Up $\uparrow$} \\
    \midrule
    1 Region  & 0\%   & 35.7M & 6.1G & 5.15 & 800K & 1.00$\times$ \\
    \midrule
    2 Regions & 64.2\% & 7.4M  & 1.4G & 7.37 & 200K & 6.30$\times$ \\
    3 Regions & 64.0\% & 7.2M  & 1.3G & 7.29 & 200K & 5.78$\times$ \\
    4 Regions & 64.0\% & 7.2M  & 1.3G & 7.55 & 200K & 5.78$\times$ \\
    \bottomrule
    \end{tabular}
    }
    \vspace{-1.2em}
\end{table}

\begin{table}[t!]
    \centering
    \caption{Effect of extending training iterations and time on an A100 GPU under a 30\% pruning rate for DDPM on CIFAR-10.}
    \label{tab:ablation_training_time_fid}
    \setlength{\tabcolsep}{16pt}

    \small 
    \resizebox{1.0\columnwidth}{!}{%
    \begin{tabular}{ccc}
        \toprule
        \textbf{Training Time (hr)} & \textbf{Training Iterations} & \textbf{FID$\downarrow$} \\
        \midrule
        3.1 & 100K & 7.95 \\
        \textbf{6.2} & \textbf{200K} & \textbf{6.89} \\
        9.3 & 300K & 7.02 \\
        12.4 & 400K & 7.19 \\
        15.5 & 500K & 7.44 \\
        \bottomrule
        \vspace{-2em}
    \end{tabular}%
    }
    
\end{table}

\section{Ablation Study on Training Iterations}
\label{sec:iterations_ablation}
To determine the most suitable training iteration under the same pruning rate, we summarize how generation quality (e.g., FID scores) changes with increased training in Table~\ref{tab:ablation_training_time_fid}. Specifically, the generation quality improves significantly, with FID scores decreasing by 1.06 when training iterations increase from 100K to 200K. Beyond 200K iterations, however, the generation quality gradually declines, suggesting an optimal training window between 100K and 200K iterations. The decline in generation quality after 200K likely reflects overfitting to specific timestep regions.

{
    \small
    \bibliographystyle{ieeenat_fullname}
    \bibliography{main}
}